\documentclass{article}

\usepackage{arxiv}
\usepackage[utf8]{inputenc} 
\usepackage[T1]{fontenc}    
\usepackage{hyperref}       
\usepackage{url}            
\usepackage{booktabs}       
\usepackage{amsfonts}       
\usepackage{nicefrac}       
\usepackage{microtype}      
\usepackage{lipsum}		
\usepackage{graphicx}
\usepackage{natbib}
\usepackage{doi}

\usepackage{algorithm}
\usepackage{algpseudocode}
\usepackage{multirow}
\usepackage{xcolor}
\usepackage[most]{tcolorbox}
\usepackage{array}
\usepackage{cancel}
\usepackage{comment}

\usepackage{array}
\usepackage{xcolor}
\usepackage[most]{tcolorbox}
\usepackage[utf8]{inputenc}
\usepackage{longtable}
\usepackage{multicol}
\usepackage{threeparttable}
\usepackage{tikz,lipsum,lmodern}
\usepackage[most]{tcolorbox}
\usepackage{graphicx}
\usepackage{subcaption}
\usepackage{xurl} 
\usepackage{hyperref} 

\newenvironment{smallerfont}{\footnotesize}{}

\definecolor{keywordcolor}{RGB}{0,0,139}
\definecolor{variablecolor}{RGB}{0,100,0}
\definecolor{green}{RGB}{0,255,0}
\definecolor{blue}{RGB}{0,0,255}
\definecolor{orange}{RGB}{255,165,0}
\definecolor{red}{RGB}{255,0,0}
\definecolor{purple}{RGB}{128,0,128}
\definecolor{cyan}{RGB}{0,255,255}
\definecolor{magenta}{RGB}{255,0,255}
\definecolor{yellow}{RGB}{255,255,0}
\definecolor{brown}{RGB}{139,69,19}
\definecolor{gray}{RGB}{128,128,128}
\definecolor{pink}{RGB}{255,182,193}
\definecolor{teal}{RGB}{0,128,128}
\definecolor{olive}{RGB}{128,128,0}
\definecolor{lightblue}{RGB}{173,216,230}
\definecolor{darkblue}{RGB}{0,0,139}
\definecolor{userpromptcolor}{RGB}{148,255,255}

\definecolor{keywordcolor}{RGB}{0,0,139}
\definecolor{variablecolor}{RGB}{0,100,0}
\definecolor{green}{RGB}{0,255,0}
\definecolor{blue}{RGB}{0,0,255}
\definecolor{orange}{RGB}{255,165,0}
\definecolor{red}{RGB}{255,0,0}
\definecolor{purple}{RGB}{128,0,128}
\definecolor{cyan}{RGB}{0,255,255}
\definecolor{magenta}{RGB}{255,0,255}
\definecolor{yellow}{RGB}{255,255,0}
\definecolor{brown}{RGB}{139,69,19}
\definecolor{gray}{RGB}{128,128,128}
\definecolor{pink}{RGB}{255,182,193}
\definecolor{teal}{RGB}{0,128,128}
\definecolor{olive}{RGB}{128,128,0}
\definecolor{lightblue}{RGB}{173,216,230}
\definecolor{darkblue}{RGB}{0,0,139}
\definecolor{userpromptcolor}{RGB}{148,255,255}

\newcommand{\chat}[2]{%
    \begin{tcolorbox}[enhanced jigsaw,breakable,boxrule=1pt,arc=4pt,outer arc=4pt,colback=blue!10!white,colframe=blue!50!black,width=6.4in,leftrule=3mm]
        \textbf{Claim:} #1 \\
        \textbf{Verification:} #2
    \end{tcolorbox}
}

\newtcolorbox{mybox}[1]{
  enhanced,
  colback=lightblue!30,
  colframe=black,
  arc=4mm,
  fonttitle=\color{white}\bfseries,
  title=#1,
  coltitle=white,
  boxrule=0.5mm
}

\newtcolorbox{userpromptbox}[1]{
  enhanced,
  colback=userpromptcolor!30,
  colframe=black,
  arc=4mm,
  fonttitle=\color{white}\bfseries,
  title=#1,
  coltitle=white,
  boxrule=0.5mm
}

\title{Towards Automated Solution Recipe Generation for Industrial Asset Management with LLM}

\author{%
    \href{https://orcid.org/0000-0002-3473-6097}{\includegraphics[scale=0.06]{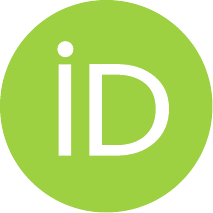}\hspace{1mm}Nianjun Zhou}\thanks{These authors contributed equally to this work.}\\
    IBM Research\\
    1101 Kitchawan Road, Route 134\\
    Yorktown Heights, NY 10598, USA\\
    \texttt{jzhou@us.ibm.com}\\
    \and
    \href{https://orcid.org/0000-0001-6210-0902}{\includegraphics[scale=0.06]{orcid.pdf}\hspace{1mm}Dhaval Patel}\footnotemark[1]\\
    IBM Research\\
    1101 Kitchawan Road, Route 134\\
    Yorktown Heights, NY 10598, USA\\
    \texttt{pateldha@us.ibm.com}\\
    \and
    \href{https://orcid.org/0009-0000-3190-3307}{\includegraphics[scale=0.06]{orcid.pdf}\hspace{1mm}Shuxin Lin}\\
    IBM Research\\
    1101 Kitchawan Road, Route 134\\
    Yorktown Heights, NY 10598, USA\\
    \texttt{shuxin.lin@ibm.com}\\
    \and
    \href{https://orcid.org/0000-0002-0275-1591}
    {\includegraphics[scale=0.06]{orcid.pdf}\hspace{1mm}Fearghal O'Donncha}\\
    IBM Research\\
    IBM Technology Campus Damastown Industrial Park\\
    Mulhuddart, Dublin, Ireland, D15 HN66\\
    \texttt{feardonn@ie.ibm.com}   
}

\renewcommand{\shorttitle}{Towards Automated Solution Recipe Generation for IAM with LLM}

\hypersetup{
pdftitle={Towards Automated Solution Recipe Generation for Industrial Asset Management with LLM},
pdfsubject={Generative AI and Industrial Asset Management},
pdfauthor={Nianjun Zhou, Dhaval Patel, Shuxin Lin, Fearghal O'Donncha},
pdfkeywords={Large Language Models, Taxonomy, Industrial Asset Management, Internet of Things,  Key Performance Indicators},
}

\begin{document}
\maketitle

\begin{abstract}
This study introduces a novel approach to Industrial Asset Management (IAM) by incorporating Conditional-Based Management (CBM) principles with the latest advancements in Large Language Models (LLMs). Our research introduces an automated model-building process, traditionally reliant on intensive collaboration between data scientists and domain experts. We present two primary innovations: a taxonomy-guided prompting generation that facilitates the automatic creation of AI solution recipes and a set of LLM pipelines designed to produce a solution recipe containing a set of artifacts composed of documents, sample data, and models for IAM. These pipelines, guided by standardized principles, enable the generation of initial solution templates for heterogeneous asset classes without direct human input, reducing reliance on extensive domain knowledge and enhancing automation. We evaluate our methodology by assessing asset health and sustainability across a spectrum of ten asset classes. Our findings illustrate the potential of LLMs and taxonomy-based LLM prompting pipelines in transforming asset management, offering a blueprint for subsequent research and development initiatives to be integrated into a rapid client solution. 

\end{abstract}


\section{Introduction}

Industrial asset management has evolved from a rigid, schedule-based approach to comprehensive practices, exemplified by frameworks like ISO 55001\footnote{https://www.iso.org/standard/55089.html}, which requires organizations to establish policies, objectives, and processes for risk management, financial management, and continuous improvement. These rigid approaches struggle to adapt to the evolving demands of modern environments. The emergence of data-rich ecosystems, pulsating with interconnected sensors and the industrial Internet of Things (IoT), necessitates a paradigm shift towards conditional asset management (CBM), a cornerstone of Industry 4.0 \cite{fernandes2021role}. An enterprise has accumulated a large amount of sensor data in its cloud system, which can be utilized to improve the efficiency and effectiveness of its maintenance strategies. A flexible maintenance framework can be established by integrating key performance indicators (KPIs) and key performance metrics related to asset health, failure scenarios, operational efficiency, and historical maintenance actions with sensor data from IoT systems, historical records, and asset profiles. This framework supports CBM to enhance financial outcomes while upholding social and environmental responsibilities. 


\begin{figure}[h]
\centering
\includegraphics[width=12cm]{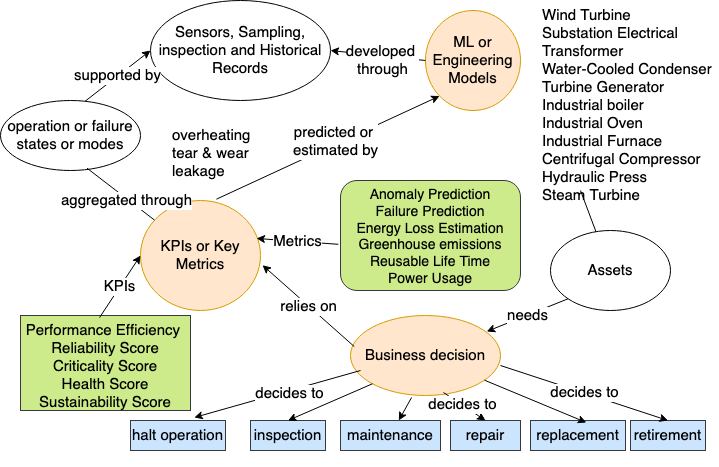}
\caption{Condition-Based Maintenance Solution Dev. Process}
\end{figure}\label{fig:conceptual_diagram}

Figure 1 presents a framework for building a CBM solution for industrial assets, such as wind turbines. This framework is an extended version of the CRISP methodology and aligns with business needs. As we can see, business users make decisions, such as repairing or inspecting, based on estimated or predicted KPIs and metrics. These KPIs are estimated, predicted, or calculated using real-time sensor data, historical records, and asset profiles with the help of data scientists or software engineers. Developing condition-based maintenance is a complex task because it involves a multi-step, multi-persona process to develop reliable models for accurately estimating or predicting KPIs and metrics. We have noticed that data scientists need a thorough educational process to understand the nature of industrial assets, analyze sensor data, and review historical records before creating a KPI-driven model solution. Presently, this process involves prolonged discussions among domain experts, data scientists, project managers, etc., making knowledge transfer time-consuming and potentially unreliable due to variations in the domain experts' knowledge depth. 


With the emergence of pre-trained Large Language Models (LLMs), we have access to a wealth of domain knowledge at our fingertips, including information about different types of industrial assets. This development opens up a promising avenue for addressing the challenges associated with CBM and positions our research at the forefront of innovation in this specialized field. Our work focuses on automatically generating a solution template, which we call a ``solution recipe'', that can be deployed within a standardized machine learning environment such as open source MLFlow\footnote{https://mlflow.org/}. A \textit{solution recipe} includes a collection of artifacts, such as domain knowledge documents, configurations, sample data, and programming codes, that are readily deployable and can be used to model and test the solution. Note that the MLFlow recipe is generated by a data scientist, and our approach is an attempt to automate that process. The most critical step in this process is to select proper IoT data, historical records, and asset profile information and convert them into meaningful KPIs and metrics, ensuring that the method is both reliable and understandable. We generate two types of solution recipes: engineering-based or multi-criteria decision analysis models that are established on recognized empirical relationships, such as health score, and machine learning-based models that include techniques like anomaly detection, failure prediction, and assessment of the expected lifespan of assets. Our research aims to automate the generation of these solution recipes, making it easier and more efficient to develop condition-based maintenance solutions.

Our research introduces two primary innovations that have the potential to significantly enhance the practical application and integration of Large Language Models (LLMs) in industrial asset management. The first innovation revolves around using taxonomy to streamline the process of prompting generation, or more specifically, taxonomy-guided prompting generation. This methodology facilitates the creation of guided prompt pipelines to produce solution artifacts and leverages a knowledge document as the foundational base for generating subsequent artifacts. Such an approach enables the automatic generation of initial solutions or solution templates tailored to various asset classes without necessitating direct interaction with domain experts.  Additionally, we have developed a validation pipeline dedicated to validating the knowledge document, ensuring the quality and accuracy of this foundational artifact with appropriate references. This validation pipeline is critical for maintaining the integrity and reliability of the entire solution-generation process.  In summary, our research unveils a guided auto-recipe generation process that automates modeling in asset management, enabling dynamic adaptation to new asset classes and client requirements for the development of advanced KPI/metric monitoring and decision-making.


%
\section{Related Work}

Our work is motivated by two streams of research: embedding domain-specific problem-solving logic into a tree-structured semantic hierarchy, and leveraging large language models (LLMs) as a guide in driving content generation for a particular aspect or topic. 


\subsection{Taxonomy-aware Prompting}
This section reviews the work in the domain of taxonomy and prompts associated with taxonomy. Taxonomy is a tree-like structure that enables end-users to organize related concepts (or terms) via relationships, which facilitates information seeking, retrieval, or behavior for downstream applications.

Existing work has largely focused on building taxonomy \cite{shah2023using, xu2023tacoprompt}, extending taxonomies \cite{xutaxoprompt, shen2020taxoexpan, liu2021temp}, and generating a taxonomy for prompt \cite{oppenlaender2023taxonomy}. These approaches, by and large, use similar mechanisms to build a large language model using data of interest and demonstrate its benefits. In our case, we aim to leverage the given taxonomy to generate appropriate prompts for extracting knowledge from LLMs. In this context, Promptiverse \cite{lee2022promptiverse} is closely related to our work, as they employed a knowledge graph to traverse and generate prompts. However, they did not utilize LLMs, and the paper's focus was on learning the knowledge graph from available information, such as videos and notes. Additionally, Promptiverse did not cover the area of writing textual inputs or instructions for prompting generative pre-trained models. Our work aims to address this gap by leveraging the given taxonomy to generate appropriate prompts for extracting domain knowledge, and utilizing LLMs to guide the content generation process as well as the prompt generation process.

In the biomedical domain, a method called HIPrompt \cite{Lu_2023} was proposed to organize domain-specific terms into a hierarchy for knowledge fusion. The objective was to use LLMs to fuse knowledge by inputting the terms in the hierarchy. However, instead of generating textual inputs to query LLMs, the authors used a ranker module to obtain the ranking of terms from the hierarchy. It's not clear why the authors chose to use a ranker module instead of generating prompts from the hierarchy. Additionally, LLMs can also be used to generate taxonomies for specific problems, such as user intent taxonomies. For example, Shah et al. \cite{shah2023using} proposed a method for generating a user intent taxonomy using LLMs. Their approach involves using LLMs to generate a set of candidate terms that represent different user intents and then using a ranker module to rank the terms based on their relevance to the target task. This approach can be helpful in situations where the hierarchy of terms is not well-defined or is too complex to be generated manually. Clearly, taxonomy is an interesting way to encode the problem, and its application using LLMs brings value to industrial applications.

\subsection{Knowledge Extraction from LLM}
This section reviews work centered around prompting techniques, multi-step reasoning, and factual verification of LLM output. LLM has been trained on the vast majority of data from Wikipedia, common-crawl books, etc. On top of this learning, the instruction version of LLM has been trained on additional demonstration to bring a logical reasoning capability on understanding user's request and responding appropriately using internal knowledge \cite{cohen2023crawling}. Such capabilities encourage communities and industries to use LLMs as Knowledge Gurus in various downstream tasks, such as the use of LLM as a web-crawler to discover factual information \cite{cohen2023crawling}.  

Extracting useful knowledge is achieved using a prompt (an instruction in natural language). LLMs are Zero-shot learners \cite{He2023, NEURIPS20228bb0d291}. Thus, we expect LLM to understand our intentions expressed via prompt accurately, but existing work often finds that its answers do not perfectly match end-user expectations. To overcome this limitation, various research techniques have been proposed to improve the logical reasoning capability, such as Chain-of-thought \cite{zhang2023automatic}, tree-of-thought \cite{yao2023tree}, etc. Moreover, to control the hallucination and to produce wrong results, Chain-of-Verification \cite{dhuliawala2023chainofverification}, Self-consistency \cite{wang2023selfconsistency}, etc. are helpful to further improve the correctness of the answer. It is important to notice that this approach does not bring any external knowledge to improve the performance of the underlying questions, such as the RAG pattern-based approach.  

Toolformer emphasizes how to use LLM as a tool to achieve tasks such as calling an external API \cite{schick2023toolformer}. External tools such as API calls, talking to knowledge graphs, etc., are made possible via several tools such as React \cite{ChaseLangChain2022}, ADaPT \cite{prasad2023adapt}, AutoGen \cite{wu2023autogen}, and. These approaches adopt a single/multiple agent system and bring a thinking capability inside LLM, along with integration with external tools, on how to solve large complex problems. Research in this area focuses on breaking down complex, high-level tasks into smaller, more manageable components. This includes techniques for reacting to and prompting certain actions, which is critical for effective task management and automation. A detailed investigation of the ADaPT paper reveals a need for task-specific demonstration techniques for a given problem, such as TextCraft, etc. How such a multi-step demonstration can be developed using taxonomy is not presented anywhere.   

The ability to self-validate the response generated by LLM depends on problem context. For example, a Question-answering system, etc, there is a ground truth that can be used to validate. On the other hand, If we generate a code, we can execute it on the sample test data to verify \cite{nori2023generalist}. However, validating the knowledge document produced by LLM is tricky. To handle the factually of generated content, WebGPT \cite{nakano2022webgpt}, WebGLM \cite{10.1145/3580305.3599931} and \cite{li2023llatrieval} are two recent approaches that demonstrated how to use LLM for generating an output which is supported by relevant reference associated with the generated output. Both papers use web as a source of ground truth while generating the response and use LLM to summarize the content with reference. In our case, we need to verify whether the generated output is factual or not. 

\section{Problem Formulation}
The problem we are addressing in our research is how to create a comprehensive solution using LLM that automatically generates a recipe for a specific industrial asset based on one Key Performance Indicator (KPI). 



\subsection{Key Terms: KPIs, Metrics and Sensors}
\begin{figure}[ht]
    \centering
    \begin{minipage}[b]{0.45\textwidth}
        \centering
        \captionof{table}{KPIs for Industrial Asset}
        \begin{tabular}{>{\raggedright\arraybackslash}p{2.5cm} >{\raggedright\arraybackslash}p{5.0cm} >{\raggedright\arraybackslash}p{2.5cm}}
        \toprule
        \textbf{KPIs} & \textbf{Meaning} \\
        \midrule
        Health Score & Reflects the current condition and performance of an asset. \\ \hline
        Sustainability Score & Evaluates the environmental sustainability of asset operations \\ \hline
        Asset Reliability & Measures the frequency of asset functioning without failure \\ \hline
        Maintenance Costs & Tracks the costs associated with maintaining an asset \\ \hline
        Asset Availability & Proportion of time an asset is operational and available for use. \\ \hline
        Performance Efficiency & Assesses the operational effectiveness of an asset. \\ \hline
        Reliability Score & Indicates the dependability of an asset over time \\ \hline
        Criticality Score & Assesses the importance of an asset to overall operations \\
        \bottomrule
        \end{tabular}
        \label{tab:KPIs}
    \end{minipage}
    \hspace{0.05\textwidth} 
    \begin{minipage}[b]{0.45\textwidth}
        \centering
        \includegraphics[width=\textwidth]{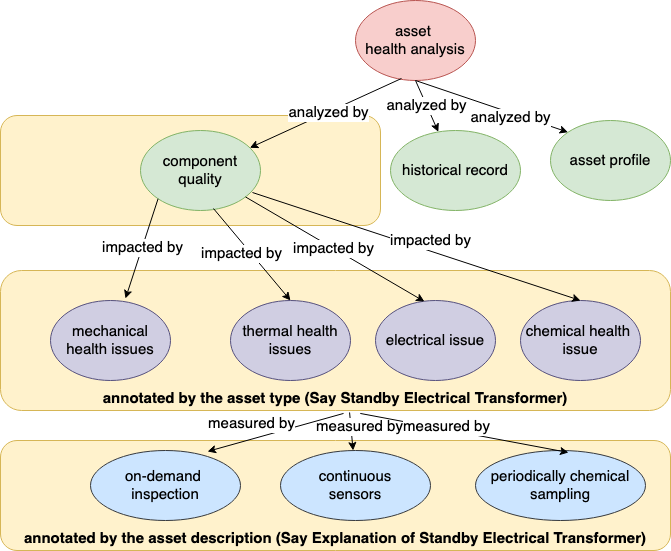}
        \caption{KPI-centric Taxonomy for Asset Health Score}\label{fig:kpitree}
        \end{minipage}
\end{figure}

Key Performance Indicators (KPIs) are the strategic measures closely linked to overarching business goals. Metrics are a broader range of operational measurements detailing various aspects of asset performance.  Both are essential for evaluating and enhancing asset performance, implementing maintenance strategies, and guiding investment decisions. The KPIs/metrics, such as mean-time-between-failure (MTBF), maintenance costs, and asset availability, provide tangible measures of an asset's operational efficiency and performance. Meanwhile, The KPIs/metrics, like failure probability, component condition monitoring, and greenhouse gas equivalent emissions, offer predictive insights, enabling the assessment of asset risk, asset health, and their environmental impact. These indicators guide informed decisions about maintenance practices, resource distribution, and long-term asset strategy formulation. Table ~\ref{tab:KPIs} lists eight KPIs commonly used in industrial asset management sectors.

IoT sensors play a critical role to estimate asset KPIs. 
For instance, vibration sensors play a pivotal role in assessing a machine's Health Score, whereas temperature sensors are crucial for maintaining performance efficiency. Similarly, energy consumption meters are instrumental in calculating metrics like Greenhouse Gas Equivalent Emissions. 
Given the diverse nature of industrial assets, including rotary equipment and fixed assets, the types of sensors required for their monitoring differ significantly. It is in this context that Large Language Models (LLMs) become valuable.

\subsection{KPI-centric Taxonomy}\label{subsec:KPI-centric_taxonomy}
A KPI-centric taxonomy is a tree-like structure that encapsulates the domain knowledge of a specific KPI. The root node of the taxonomy tree is the KPI, while the child nodes are factors that influence the KPI. The leaf nodes are either factors or sensors, annotated by asset types. The edges represent the relationships between the parent and child nodes.

For instance, Figure \ref{fig:kpitree} shows a taxonomy for the asset health KPI, where the health of an asset is analyzed by component quality, etc. By traversing the taxonomy from top to bottom, the domain knowledge of asset health for an industrial asset can be obtained. It's important to note that the taxonomy doesn't include asset-specific information, as LLM helps generate this knowledge.

Although we assume that the KPI-centric taxonomy is available as input, it's not necessary to assume that it has captured all the information. Overall, taxonomy allows domain experts to  contribute their specialized knowledge, which is vital for contextualizing the solution within its specific industrial setting.


\subsection{Solution Recipe}
To support the function of predicting/estimating a KPI for an industrial asset, we generate a solution entity named \textit{recipe}. For an industrial asset class, a \textit{recipe} is a collection of documentation and programming codes denoted by \(R = \{a_1, a_2, \ldots, a_n\}\). Each recipe $R$ is designed to handle a specific KPI and metric. Currently, we break down a recipe into five artifacts, including:

\begin{itemize}
    \item \textbf{Knowledge Document for KPI/metric Analysis}: is a markdown file that provides a comprehensive introduction to the asset, including its definition and a detailed overview of the factors and sensors or another measurement such as inspection, sampling, and historical record and asset profile that could affect the KPI/metric estimation or forecasting.  
    \item \textbf{KPI/metric Modeling Configuration(s)}: one or more YAML file(s) that specifies the detailed information regarding each sensor and historical records, including description, unit if required, and value range when the value is in a normal or abnormal state or situation, etc. and the assignment of the scale to the value.
    \item \textbf{Sample Dataset}: Typically, it is a CSV (comma-separated value) dataset including identified sensor time series, variables from historical records, or asset profiles. They can be real-world industry data or synthetic data generated based on a simulated program (itself could be parts of the recipe). 
    \item \textbf{Modeling Code}: programming classes or functions that predict or estimate a KPI/metric from sensors' or other monitored data and historical or profile information. These models can be either data-driven or traditional engineering models.
    \item \textbf{Recipe Wrapper}: is an artifact that encapsulates the code for integrating inputs from both configuration settings and the dataset, and subsequently invoking the model code to facilitate model generation. 
\end{itemize}

Large Language Models (LLM), such as ChatGPT, LLama, Mixtral, and Granite, are trained on vast multi-domain textual data curated from various sources, including books, literature, and texts gathered from online resources. These models, with their multifaceted knowledge, are not only capable of providing domain-specific insights but also generating programming code based on user instructions. In this paper, we present a solution that leverages LLM and KPI-centric taxonomy to automatically generate a \textit{recipe}.

\section{Solution Outline}
\begin{figure}[ht]
    \centering
    \begin{minipage}[b]{0.45\textwidth}
        \centering
        \includegraphics[width=\textwidth]{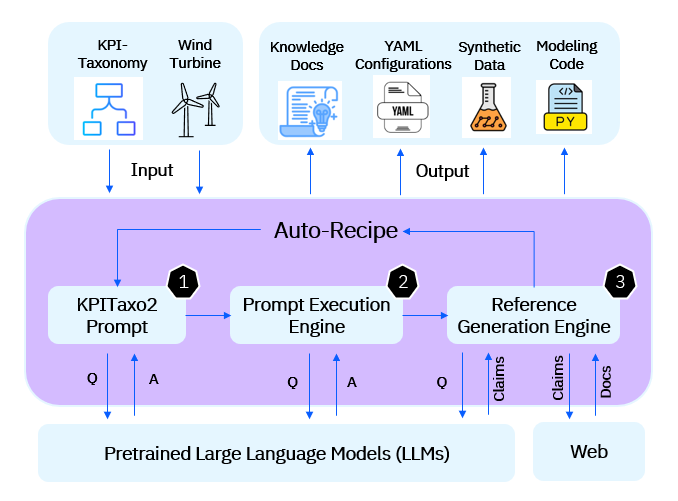}
        \caption{Architecture Diagram for AutoRecipe Generation}
        \label{fig:architecture_diagram}
        \end{minipage}
    \hspace{0.05\textwidth} 
    \begin{minipage}[b]{0.45\textwidth}
        \centering
        \includegraphics[width=\textwidth]{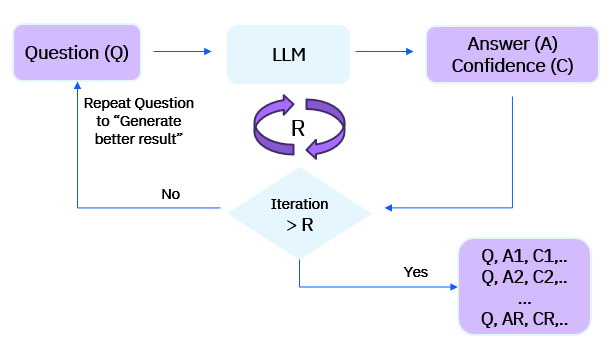}
        \caption{Iterative Flow for Quality Refinement}
        \label{fig:iterative}
        \end{minipage}
\end{figure}

This work proposes a framework for AutoRecipe, an automated solution for creating a Solution Recipe for an industrial asset. Figure ~\ref{fig:architecture_diagram} illustrates the system architecture. 

The first crucial task in AutoRecipe is to generate a series of meaningful prompts, hereinafter referred to as PromptSequence, using a KPI-centric taxonomy. The prompt constitutes a critical element in the architecture of a high-performing Large Language Model (LLM), yet the organization of prompts into sequences also demands careful consideration. This aspect has been discussed by research into methodologies like chain-of-thought, tree-of-thought, and graph-of-thought, underscoring the significance of structured prompt sequences in enhancing LLM performance \cite{wu-etal-2023-chain, yao2023tree, besta2023graph}.  In Section \ref{kpip}, we discuss the KPITaxo2Promt methodology, a two-step hierarchical planning-aware ADaPT approach to generate a variable-length PromptSequence employing LLM. 

Once domain-specific prompts are discovered, they are executed in a particular order. Section \ref{psee} offers five types of prompt execution methodologies, including ReACT~\cite{ChaseLangChain2022}. Overall, we have designed three distinct pipelines to generate artifacts for a recipe, with each pipeline tailored to transition smoothly from domain knowledge generation to a deployable solution. Finally, we propose a qualitative validation of the generated artifacts. The question remains: how can we trust the outcomes produced by LLMs, given their propensity for generating hallucinations? We tackle this challenge from two angles: 1). Ensuring LLMs generate output with a higher confidence in the output quality, and 2) leveraging external knowledge, such as the web, to validate the factuality of the content. Section~\ref{qagr} discusses the details.

\begin{mybox}{KPITaxo2Prompt: Generate PromptSequence}
\label{prompt:kpitaxo}
  \small
  \textbf{Task Description:} Come up with a short plan ...

  \vspace{1pt} 

  \textbf{Taxonomy Intro:} [\textit{parent}] is [\textit{relation}] by [\textit{child}] ...

  \vspace{1pt} 

  \textbf{Materialized Taxonomy:} Here is the asset health taxonomy.

  \begin{itemize}
    \item Asset health is the root node.
    \item Asset health is analyzed by \fbox{component}.
    \item Asset health is analyzed by historical record.
    \item ...
  \end{itemize}

  \vspace{2pt} 

  \textbf{Goal:} Calculate asset health using \fbox{component}.

  \vspace{1pt} 

  \textbf{Think:} My target is the \fbox{component}. ...

  \begin{enumerate}
    \item \textit{Step 1:} Let us focus on the \fbox{component}-based asset health. ... factors coming from the 1. \colorbox{cyan}{Mechanical}, 2. \colorbox{pink}{Electrical} ...
  \end{enumerate}

    \textbf{Think:} Now, I will traverse the taxonomy for each child node ...

  \begin{enumerate}
    \item \textit{Step 2:} Let us focus on \colorbox{cyan}{Mechanical} issue ...
    \item \textit{Step 3:} Let us focus on \colorbox{pink}{Electrical} issue ...
    \item ...
  \end{enumerate}

  \textbf{Think:} ... Let us be more specific for an {asset class}.

  \begin{enumerate}
    \item \textit{Step X:} ...
  \end{enumerate}

  \textit{Step Y:} Generate a Python code that: ...
  ...
\end{mybox}

\subsection{KPITaxo2Prompt Generation}
\label{kpip}
In the proposed prompt \hyperref[prompt:kpitaxo]{\textcolor{blue}{KPITaxo2Prompt}}, we aim to utilize an instruction-tuned LLM to generate a prompt sequence that supports KPI-centric taxonomy use cases. Building upon the ADaPT approach~\cite{prasad2023adapt}, our proposed prompt includes an expanded version of the \textbf{taxonomy introduction} section, which lays the foundation for a context-aware taxonomy. Additionally, we have included a \textbf{materialized taxonomy} (specifically, asset health) to provide a concrete example. Unlike ADaPT, our approach does not rely on a fixed goal that can be evaluated each time. Instead, we have implemented a planner that comprehends the taxonomy and generates a comprehensive plan. Our goal is to calculate asset health using components, just one example of how our proposed prompt can be applied.

The remainder of the prompt consists of a sequence of Think-Step(s)-Think cycles, where we demonstrate how the LLM can generate prompt sequence \colorbox{pink}{Step 1} $\rightarrow$
\colorbox{pink}{Step 2}
$\rightarrow$
\colorbox{pink}{Step 3}
$\rightarrow$
...
$\rightarrow$
\colorbox{pink}{Step X}
$\rightarrow$
\colorbox{pink}{Step Y}. In each cycle, the Think step serves as information collectors from the taxonomy, while the \textit{Step P} steps involve the LLM generating question/code which will be part of PromptSequence. 



In our approach, we adopted three principles to guide the generation of steps: (i) starting with generic information and gradually adding specific details (compare \colorbox{pink}{Step 1} vs \colorbox{pink}{Step Y}); (ii) capturing forward-looking information in questions using parent-child relationships inferred from the taxonomy (see \colorbox{pink}{Step 1}); and (iii) repeating questions to improve quality. By following these principles, we can generate an effective interrogation of the entire taxonomy. However, we noticed a weakness in the LLM's reasoning ability to generate conditional questions that require accounting for distant parent-child nodes of the taxonomy. To overcome this limitation, we leveraged the AI4Code approach, where questions that require conditional evaluation are generated as Python code first (see \colorbox{pink}{Step Y}), which are later converted into questions in a second round of execution. Essentially, KPITaxo2Prompt is a two-step approach where step 1 generates a sequence of prompts, some of which may be code, and the execution of these prompts in the second round generates the final questions.

\subsection{PromptSequence Execution Engine}
\label{psee}
Given the PromptSequence, which is a guided pipeline, the first task we execute is to generate the knowledge document. We defined five distinct approaches to generate outputs using PromptSequence:
\begin{itemize}
\item Last Question ($Last_Q$): Only the last question in PromptSequence is executed to capture the Zero-shot capability of LLM as a baseline.
\item All Questions Concatenated ($All_Q$): We combined all questions from PromptSequence into one extended query, testing the effect of presenting the full question context in a single prompt. 
\item All Questions with Chain of Thought ($All_{QCOT}$): This method enhances the $All_Q$ approach by incorporating a ``think step-by-step'' comment at the end of last questions in the prompt.  
\item All Questions with ReAct ($All_{QREACT}$): This method enhances the $All_Q$ approach by incorporating a ReAct agent that enables LLM think-act-observe each questions in PromptSequence before answering the last question.
\item Guided Iterative Thought ($GIT_Q$): This approach simulates a dynamic Q/A session, where each question and its subsequent answer lead to the next query, mirroring a real-world interaction pattern.  In this process, questions are already generated in PromptSequence and its answers help to understand how the final knowledge is generated.
\end{itemize}


These methods aim to optimize the extraction of domain-specific knowledge for a particular KPI, evaluating their effectiveness in contributing to a rich, accurate knowledge base. In the context of Asset Health KPIs for Wind Turbine, the expected content of the knowledge document discusses the wind turbine's components, specific sensors, and their corresponding health monitoring responsibilities. This document also establishes criteria for health status labeling across diverse sensor outputs, categorizing them into distinct classes, such as normal, warning, or critical. This initial stage lays the groundwork for a tailored monitoring framework, precisely tailored to the wind turbine's unique requirements.






\subsection{Quality Assurance of Generated Recipe}
\label{qagr}
As the knowledge document is the base for the generation of artifacts, we employ a continuous booster mechanism to improve the knowledge document. This process, iterated \textit{N} times, is designed to enhance the quality of results and incrementally increase the content tied to the knowledge extraction outputs. We incorporate factual scores alongside references to external sources, thereby grounding our solutions in verifiable information. This method of including confidence from external resources, with references linked directly within our outputs, ensures an increased degree of accuracy and reliability. 

\subsubsection{Iterative Information Discovery Flow}\label{subsubsec:iteractive}

In this subsection, we introduce a technique called Iterative Information Discovery Flow to enhance the information discovery process using LLM (See Figure \ref{fig:iterative}). We leverage LLM's built-in capability to generate a confidence score (C) for the output. For each question (Q) we pose to LLM, we not only produce an answer (A) but also a confidence score ranging from 0 to 100 percent. The process iterates up to ten times, and after each round, we evaluate the quality of the generated text. The question and answer from the current round are appended to the System Prompt of the agent, and the original question Q is supplemented with the additional text: "generate a better result." Unlike LLRetrival, we don't rely on external sources to calculate the score. If the confidence score of the generated text in the current round falls below a certain threshold, we terminate the process to avoid repetition. Note that, confidence score act as quality score and if quality score is poor the answer is dropped. 

\subsubsection{Reference Generator Pipeline}\label{subsubsec:reference_generator_pipeline}
Our task is formalized as follows: Given a knowledge document ($kd$) and a corpus of extracted text passages from web $D$, the pipeline is required to return an output $S$, which consists of $n$ passages, $p_1, p_2, ..., p_n$, and each passage $p_i$ cites a list of passage $D_i$ = \{$d_{i,1}$, $d_{i,2}$, ..., $d_{i,N}$\}; where $d_{i,1}$ $\in$ $D$. In this work, we segment LLMs output into passages by parts in knowledge documents. While LLMs output may include tables, numbers, etc, we noticed that all the passages have valuable information and need citation. In this work, we generate citations per part, not by sentences.

Note that, we do not have a web corpus $D$ available upfront. Thus, one objective is to construct it for a $kd$ selectively. As we have divided the knowledge document ($kd$) into passages $p_i$, we use the instruction version of LLM to generate the three sub-claims $claim_i^1$, $claim_i^2$ and $claim_i^3$. We have used a demonstration approach as in-context learning approach to teach LLM. Then, each claim $claim_i^1$ is used as a web query to fetch relevant web-URLs and, subsequently a web documents. We used open-source DDGS, a popular search engine, to collect the claim-centric data passages. Then, we use TRUE, a T5-11B model fine-tuned on a collection of NLI datasets, to identify web-passage entails the sub-claims. As TRUE targets the factual correctness and has been demonstrated a valuable tool, AutoRecipe use it automatically generating evidence available on internet for each passage. 
     


\section{Experiment}

To test our approach, we first created a taxonomy for two key performance indicators (KPIs): Asset Health and Sustainability. Then, we used Asset Health to prepare a KPITaxo2Prompt, as discussed in Section 3. To validate that KPITaxo2Prompt would work for the Sustainability KPI, we generated a prompt sequence without making any modifications. As expected, we found that just one internal thought was enough to transfer knowledge from one demonstration to another, confirming our initial hypothesis of "write once and apply multiple times" on different KPIs. Finally, we used the generated prompt sequences to create knowledge documents for 10 assets.

\subsection{Knowledge Extraction Improvement}\label{subsec:knowledge_extraction_experiment}
\begin{figure}[ht]
    \centering
    \begin{minipage}[b]{0.45\textwidth}
        \centering
        \includegraphics[width=\textwidth]{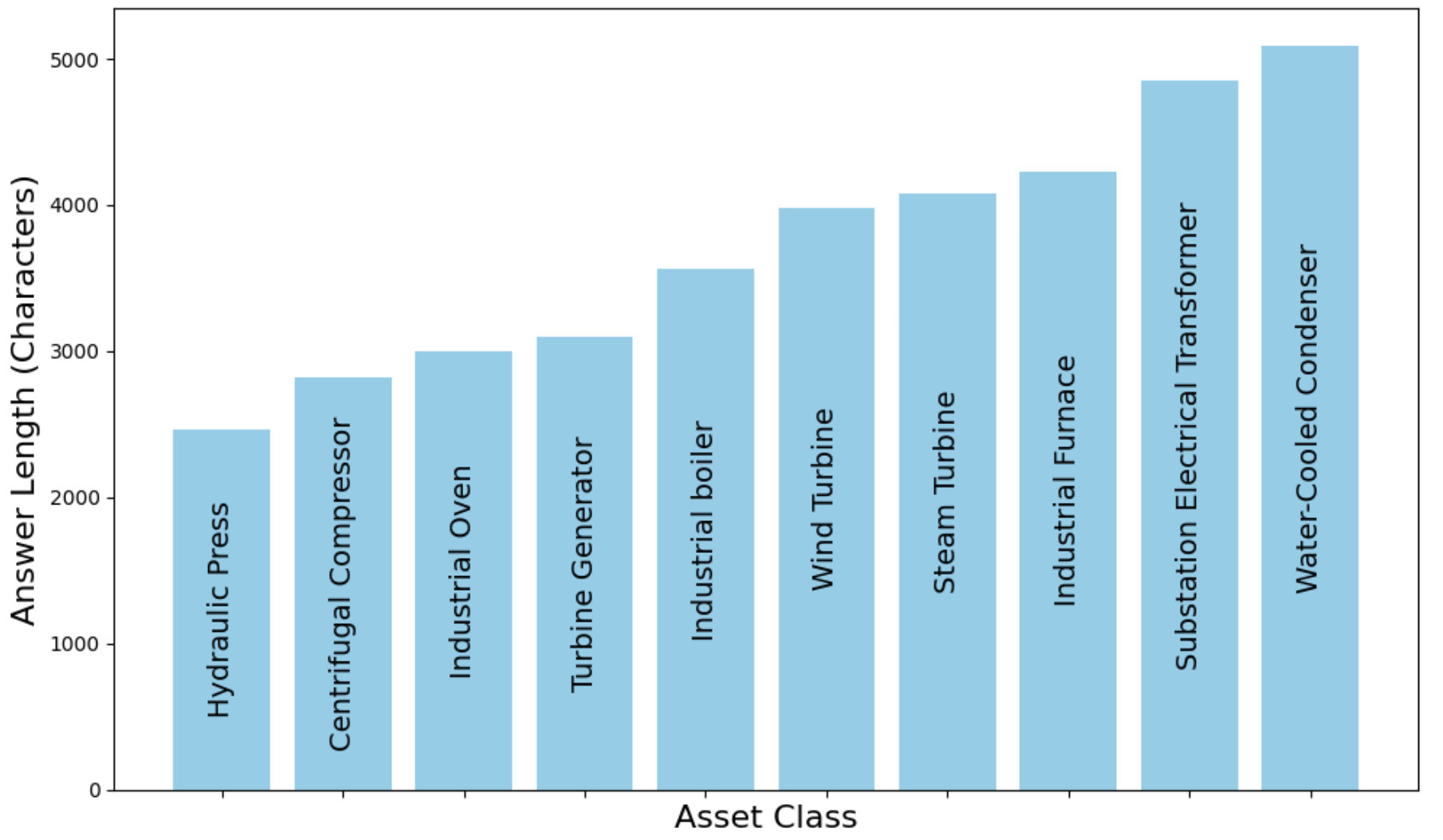}
        \caption{Answer Length by Asset Class}
        \label{fig:exp1}
        \end{minipage}
    \hspace{0.05\textwidth} 
    \begin{minipage}[b]{0.45\textwidth}
        \centering
        \includegraphics[width=\textwidth]{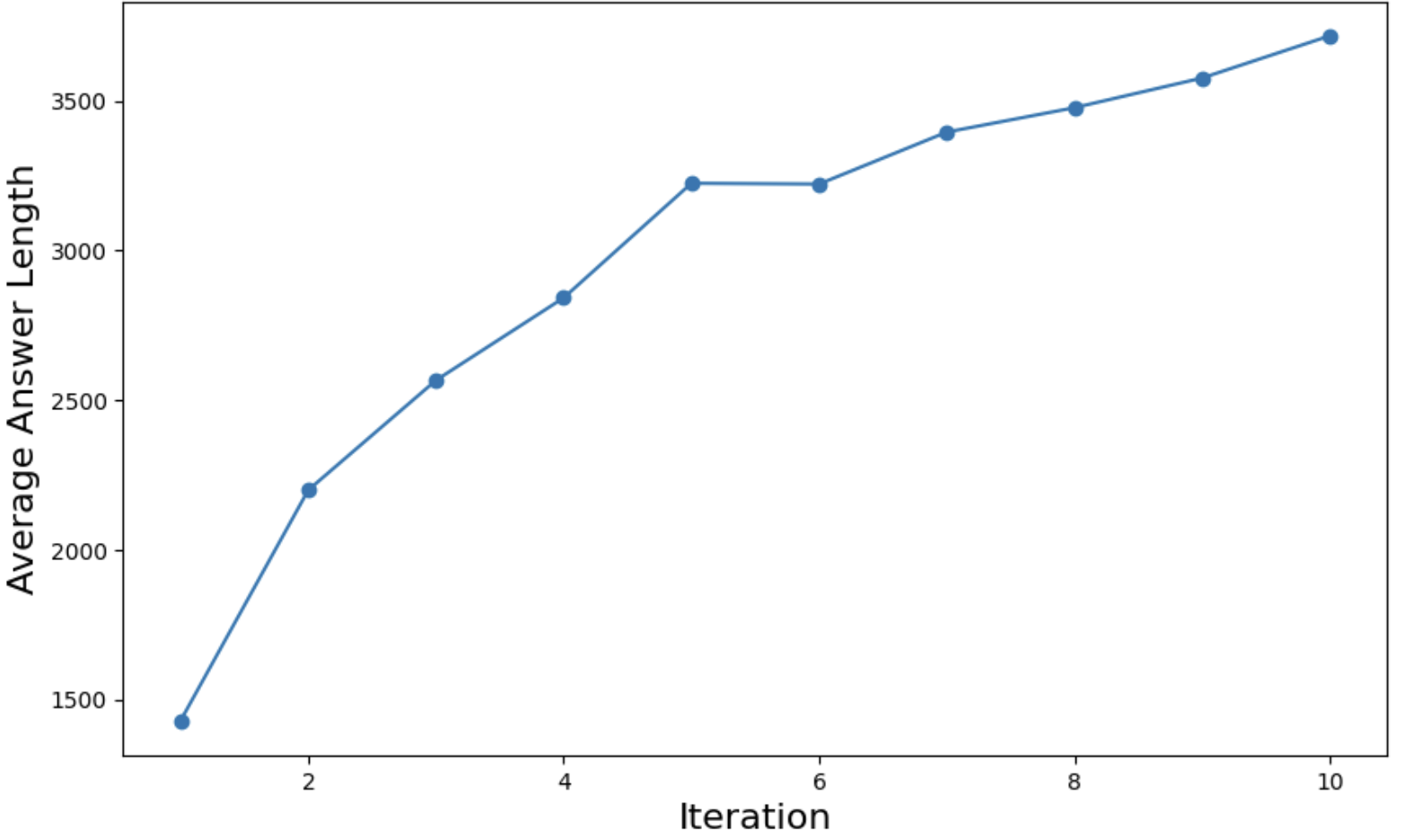}
        \caption{Answer Length by Iteration}
        \label{fig:exp2}      
        \end{minipage}
\end{figure}

To demonstrate how the technique of Section \ref{subsubsec:iteractive} works, we conducted a small-scale experiment involving ten asset classes to obtain asset descriptions as knowledge. Figure \ref{fig:exp1} shows the answer length obtained in the last iteration for each asset class. We can see that the answer length varies across asset classes. We recorded how these answer lengths vary over different iterations in Figure \ref{fig:exp2}. The incremental trend of the document length over the iterations shows the effectiveness of reinforcement in helping complete the knowledge extraction. We observed that in most cases, the second iteration significantly increased the content length compared to the remaining rounds.

\begin{figure}[ht]
    \centering
    \begin{minipage}[b]{0.45\textwidth}
        \centering
        \includegraphics[width=\textwidth]{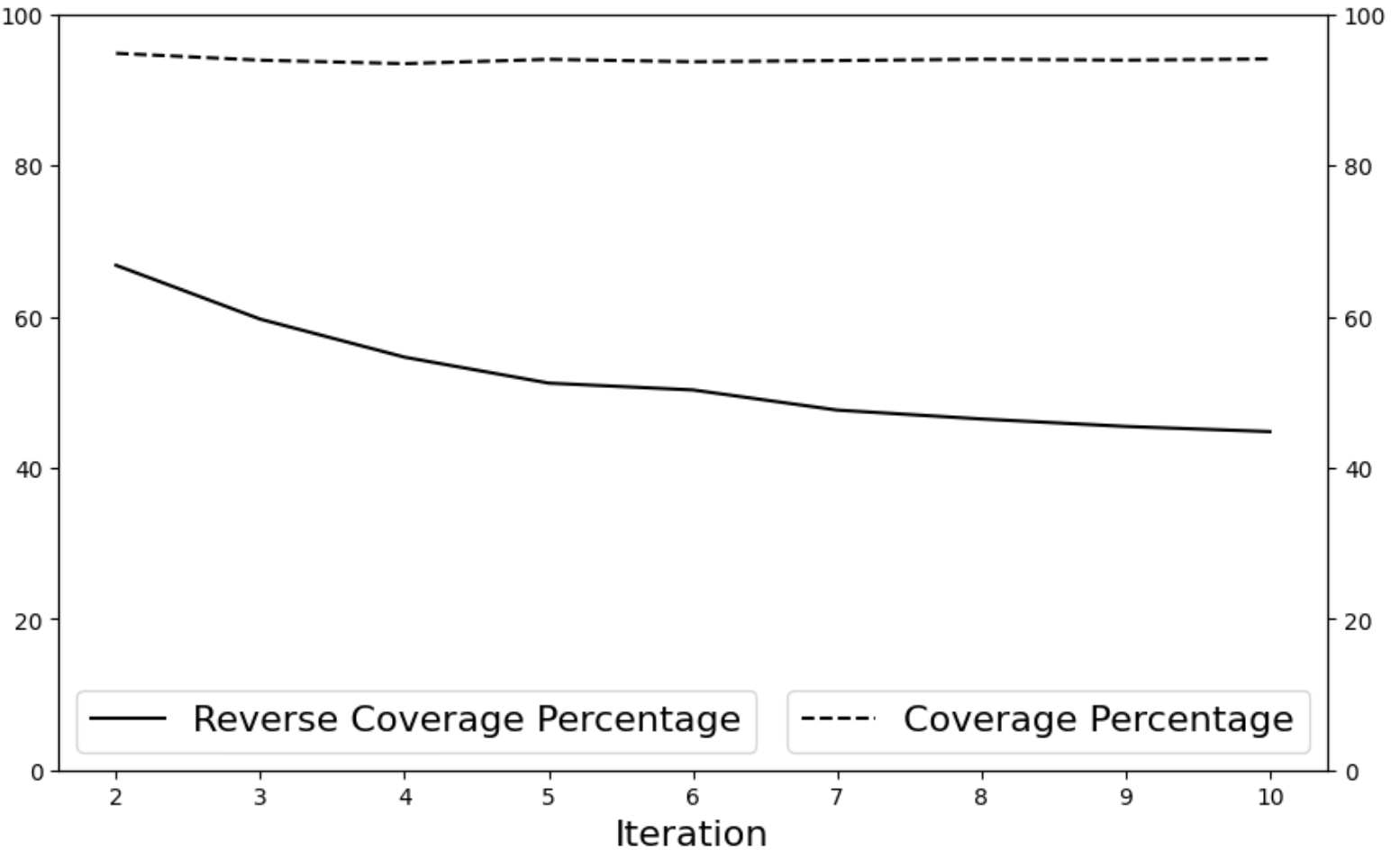}
        \caption{Averaged Coverage Percentages}
        \label{fig:exp3}
        \end{minipage}
    \hspace{0.05\textwidth} 
    \begin{minipage}[b]{0.45\textwidth}
        \centering
        \includegraphics[width=\textwidth]{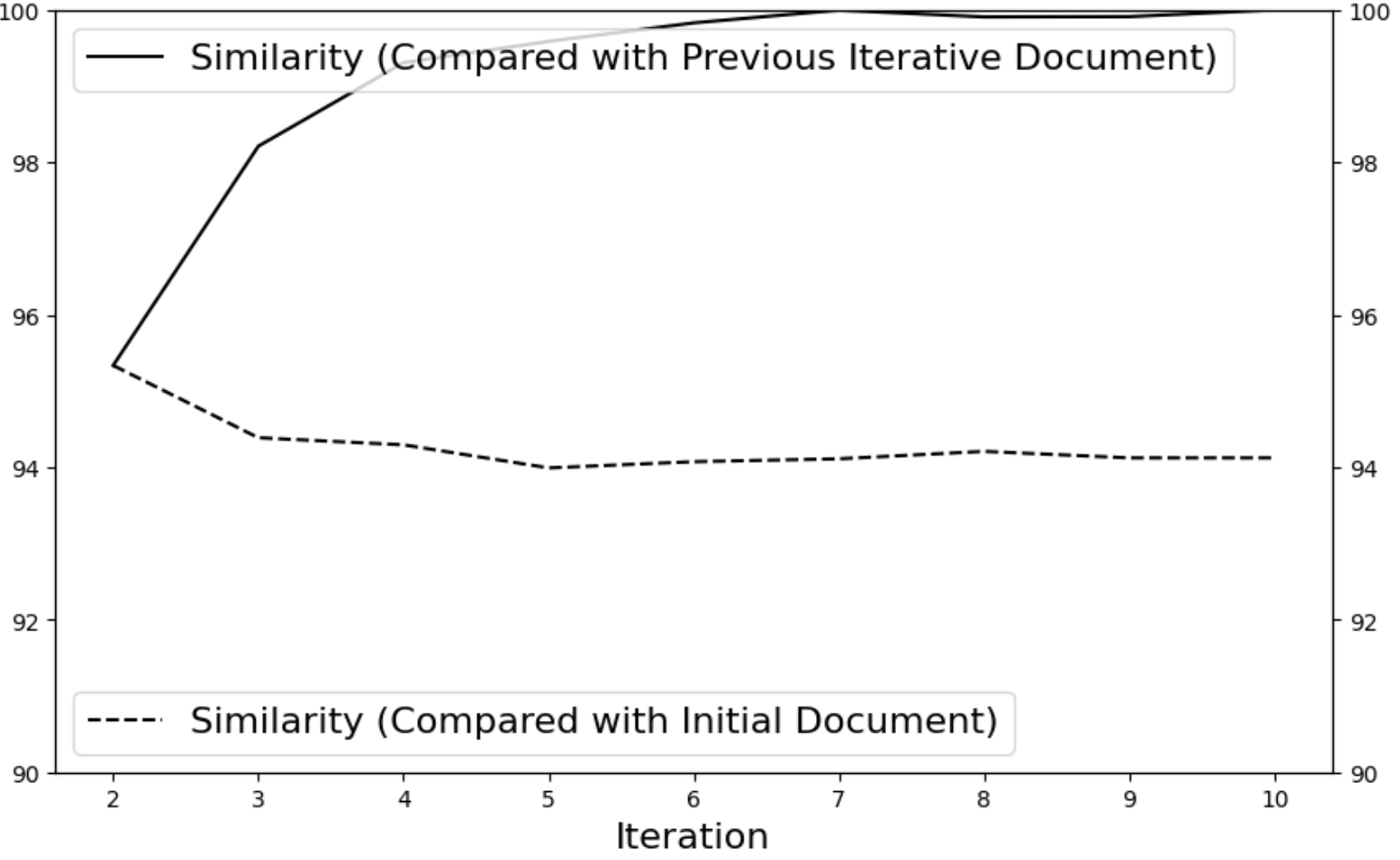}
        \caption{Averaged Semantic Similarity Scores}
        \label{fig:exp4}
        \end{minipage}
\end{figure}

We also conducted coverage analysis and a semantic study of the generated documents over iterations. We used the \textit{CountVectorizer} from \textit{scikit-learn} for the coverage analysis. Content-wise, Figure \ref{fig:exp3} shows that the iterative documents cover the initial document. However, the coverage from the initially generated document to the iterative generated documents decreased over the iterations. The semantic similarity score is symmetric when comparing two documents. Figure \ref{fig:exp4} shows that the semantic contents interactively increase and become saturated by the eighth iteration.

Overall, the study analyzes how iterative prompting affects answer length, coverage, and semantic similarity. The results suggest that as the number of iterations increases, the knowledge document becomes longer, but the semantic meaning and coverage become stable, indicating that the quality improves and reaches saturation in around seven iterations.

\begin{figure*}[!]
\centering
\frame{\includegraphics[width=14cm]{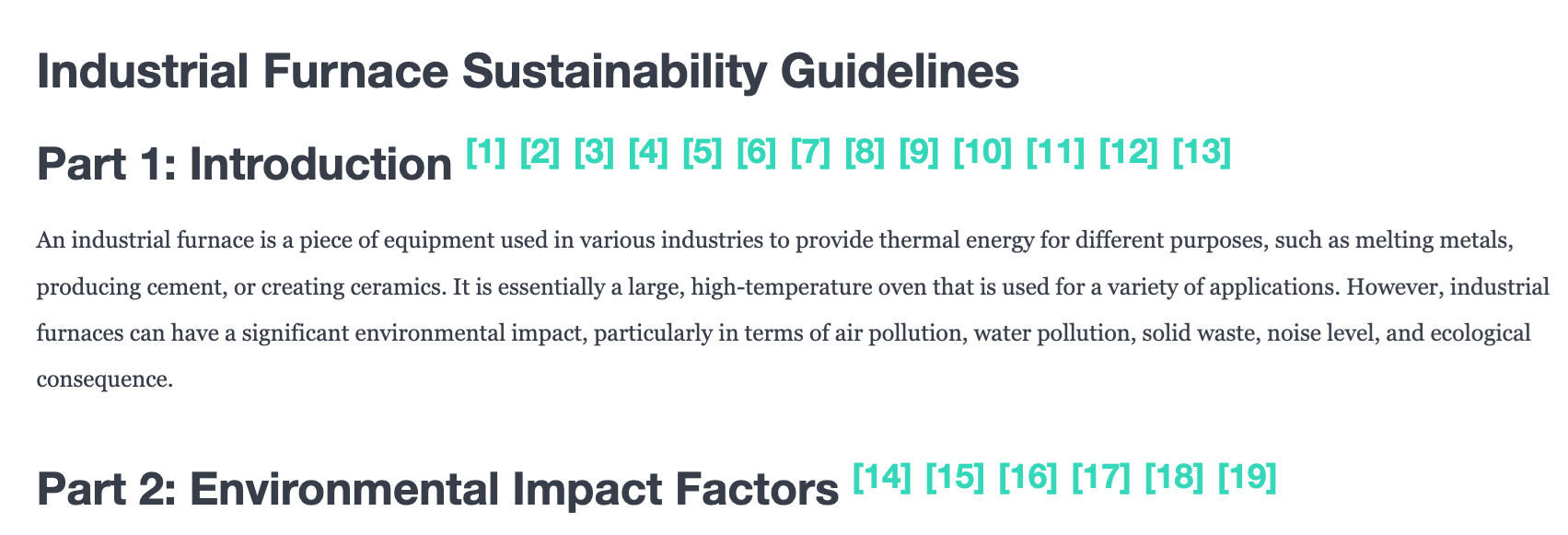}}
\caption{Knowledge Document for Sustainability Analysis with Reference (Example: Industrial Furnace)}
\label{fig:kddoc}
\end{figure*}

\subsection{Knowledge Extraction Factual Validation}

We apply \textit{Reference Generator Pipeline} (See Section~\ref{subsubsec:reference_generator_pipeline}) on two asset classes and KPIs (See Table~\ref{tab:verification_experiments}). Since, the knowledge document is spitted into three parts (asset description, KPI specific explanation, and measurement approaches), we tracked the claims at part level. For example, (3,2,3) for Wind turbine's sustainability score indicates 3, 2 and 3 claims are generated by our LLM based claim generator for each part respectively. The last columns indicates unique number of URLs collected by AutoRecipe for each passage. Once reference generation step is concluded, the knowledge document will be enriched with the external knowledge as shown in Figure \ref{fig:kddoc}.  

\begin{table}[ht]
\centering
\caption{Reference Generation for Knowledge Documents}
\label{tab:verification_experiments}
\small
\begin{tabular}{c|ccl} 
\toprule
Asset  & KPI & Claims & Web URLs Identified \\
Class &  & Generated & Identified \\ 
\midrule
Wind  & Health & (3,3,3) & ( 53,107,149) \\ \cline{2-4}
Turbine & Sustainability & (3,2,3) & ( 99, 72, 72) \\ \hline
Industrial & Health & (3,3,3) & (104,111,129) \\ \cline{2-4}
Furnace & Sustainability & (3,3,3) & ( 92,146,123) \\ 
\bottomrule
\end{tabular}
\end{table}

As an illustration, here is the two exemplary  claims for \textit{wind turbine} asset health, which are used to fetch the web urls for reference documents. We have also added a verification results that are obtained using NLI pre-trained model. Table \ref{tab:unique_urls} display the collected URL which are already validated. The selection of URL is based on the search engine and more powerful search engine, the better result and thus better references.

\chat{\small A (Generated from Part 1: Asset Description). A wind turbine converts the kinetic energy of wind into electrical energy by first converting it into mechanical energy.}{True}

\chat{\small B (Generated from Part 3: Component Condition Measurement using Sensors). Oil analysis sensors, vibration sensors, temperature sensors, and displacement sensors can be used to monitor the condition of a wind turbine gearbox and main bearing and can help identify issues with the mechanical components and thermal performance of the equipment.}{True}

\begin{table}[ht]
\small
\centering
\caption{Subset of Unique URLs validated by NLI system for each passage} 
\label{tab:unique_urls}
\begin{tabular}{p{1cm}p{13.0cm}} 
\toprule
\textbf{Passage ID} & \textbf{Unique URLs} \\
\midrule
1 & \url{https://abbottfurnace.com/a-guide-to-industrial-furnace-features-and-types/} \\
  & \url{https://www.sciencedirect.com/topics/engineering/industrial-furnace}\\
\midrule
2 & \url{https://www.sciencedirect.com/science/article/pii/S0957582021000549} \\
\midrule
3 & \url{https://www.mdpi.com/1424-8220/20/11/3113} \\
\bottomrule
\end{tabular}
\end{table}

\subsection{Pipeline Execution Engine}
\begin{figure}[h!]
\centering
\includegraphics[width=12cm]{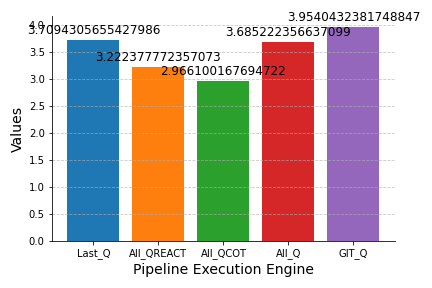}
\caption{Comparison of Various Execution strategy for wind turbine asset health knowledge document}
\label{fig:exp5}
\end{figure}

We also applied the all 5 pipelines to obtain the different depth of documents. We use popular TTR (text to token ration - which is the ratio of the number of unique types to the square root of the total number of tokens) to measure the obtain the first level of understanding for generated document. The high value of TTR is preferred. Table \ref{fig:exp5} shows the comparison among existing Prompt method.  

\subsection{Appendix Summary}
Appendix \ref{sec:promt_design} discuss various prompt used as a part of experimental work. We have used Myxtral7x13b as a LLM for all the experimental work. The Sustainability KPI taxonomy is listed in Appendix \ref{kctax}. The entire system prompt to generate the prompt sequence from taxonomy is given in Appendix \ref{prompt:kpitaxo_sus} including prompt sequence. The sample react prompt for pipeline $ALL_{QREACT}$ is given in Appendinx \ref{pmq}. The generated knowledge doc is provided in Figure \ref{fig:knowledgedoc} and \ref{fig:ref1}.

\section{Conclusion}
Our solution and experiment have systematically explored the capability of LLMs to generate detailed and relevant solution recipes (docs, sample data, and codes) across a spectrum of industrial asset classes through a structured approach using tailored prompts that could be even auto-created using KPI-centric taxonomy. By testing the LLM's performance across ten asset classes, we aimed to validate the model's consistency and completeness in knowledge extraction, which is critical for the quality of our knowledge base. Our methodology, employing a blend of system and user prompts as detailed in the Appendix, was designed to not only capture the essence of each asset class but also to ensure the adaptability and precision of the LLM in addressing domain-specific inquiries.

\bibliographystyle{unsrtnat}


\begin{thebibliography}{27}
\providecommand{\natexlab}[1]{#1}
\providecommand{\url}[1]{\texttt{#1}}
\expandafter\ifx\csname urlstyle\endcsname\relax
  \providecommand{\doi}[1]{doi: #1}\else
  \providecommand{\doi}{doi: \begingroup \urlstyle{rm}\Url}\fi

\bibitem[Fernandes et~al.(2021)Fernandes, Reis, Mel{\~a}o, Teixeira, and Amorim]{fernandes2021role}
Jorge Fernandes, Jo{\~a}o Reis, Nuno Mel{\~a}o, Leonor Teixeira, and Marlene Amorim.
\newblock The role of industry 4.0 and bpmn in the arise of condition-based and predictive maintenance: A case study in the automotive industry.
\newblock \emph{Applied Sciences}, 11\penalty0 (8):\penalty0 3438, 2021.

\bibitem[Shah et~al.(2023)Shah, White, Andersen, Buscher, Counts, Das, Montazer, Manivannan, Neville, Ni, et~al.]{shah2023using}
Chirag Shah, Ryen~W White, Reid Andersen, Georg Buscher, Scott Counts, Sarkar Snigdha~Sarathi Das, Ali Montazer, Sathish Manivannan, Jennifer Neville, Xiaochuan Ni, et~al.
\newblock Using large language models to generate, validate, and apply user intent taxonomies.
\newblock \emph{arXiv preprint arXiv:2309.13063}, 2023.

\bibitem[Xu et~al.(2023)Xu, Liu, Niu, Chen, Cai, Wen, and Yuan]{xu2023tacoprompt}
Hongyuan Xu, Ciyi Liu, Yuhang Niu, Yunong Chen, Xiangrui Cai, Yanlong Wen, and Xiaojie Yuan.
\newblock Tacoprompt: A collaborative multi-task prompt learning method for self-supervised taxonomy completion.
\newblock In \emph{Proceedings of the 2023 Conference on Empirical Methods in Natural Language Processing}, pages 15804--15817, 2023.

\bibitem[Xu et~al.(2022)Xu, Chen, Liu, Wen, and Yuan]{xutaxoprompt}
Hongyuan Xu, Yunong Chen, Zichen Liu, Yanlong Wen, and Xiaojie Yuan.
\newblock Taxoprompt: A prompt-based generation method with taxonomic context for self-supervised taxonomy expansion.
\newblock pages 4432--4438, 2022.

\bibitem[Shen et~al.(2020)Shen, Shen, Xiong, Wang, Wang, and Han]{shen2020taxoexpan}
Jiaming Shen, Zhihong Shen, Chenyan Xiong, Chi Wang, Kuansan Wang, and Jiawei Han.
\newblock Taxoexpan: Self-supervised taxonomy expansion with position-enhanced graph neural network.
\newblock In \emph{Proceedings of The Web Conference 2020}, pages 486--497, 2020.

\bibitem[Liu et~al.(2021)Liu, Xu, Wen, Jiang, Wu, and Yuan]{liu2021temp}
Zichen Liu, Hongyuan Xu, Yanlong Wen, Ning Jiang, Haiying Wu, and Xiaojie Yuan.
\newblock Temp: taxonomy expansion with dynamic margin loss through taxonomy-paths.
\newblock In \emph{Proceedings of the 2021 Conference on Empirical Methods in Natural Language Processing}, pages 3854--3863, 2021.

\bibitem[Oppenlaender(2023)]{oppenlaender2023taxonomy}
Jonas Oppenlaender.
\newblock A taxonomy of prompt modifiers for text-to-image generation.
\newblock \emph{Behaviour \& Information Technology}, pages 1--14, 2023.

\bibitem[Lee et~al.(2022)Lee, Chung, Kim, Song, and Kim]{lee2022promptiverse}
Yoonjoo Lee, John Joon~Young Chung, Tae~Soo Kim, Jean~Y Song, and Juho Kim.
\newblock Promptiverse: Scalable generation of scaffolding prompts through human-ai hybrid knowledge graph annotation.
\newblock In \emph{Proceedings of the 2022 CHI Conference on Human Factors in Computing Systems}, pages 1--18, 2022.

\bibitem[Lu et~al.(2023)Lu, Shen, Xiong, Ma, Staab, and Yang]{Lu_2023}
Jiaying Lu, Jiaming Shen, Bo~Xiong, Wenjing Ma, Steffen Staab, and Carl Yang.
\newblock Hiprompt: Few-shot biomedical knowledge fusion via hierarchy-oriented prompting.
\newblock In \emph{Proceedings of the 46th International ACM SIGIR Conference on Research and Development in Information Retrieval}, SIGIR ’23. ACM, July 2023.
\newblock \doi{10.1145/3539618.3591997}.
\newblock URL \url{http://dx.doi.org/10.1145/3539618.3591997}.

\bibitem[Cohen et~al.(2023)Cohen, Geva, Berant, and Globerson]{cohen2023crawling}
Roi Cohen, Mor Geva, Jonathan Berant, and Amir Globerson.
\newblock Crawling the internal knowledge-base of language models, 2023.

\bibitem[He et~al.(2023)He, Xie, Jha, Steck, Liang, Feng, Majumder, Kallus, and Mcauley]{He2023}
Zhankui He, Zhouhang Xie, Rahul Jha, Harald Steck, Dawen Liang, Yesu Feng, Bodhisattwa~Prasad Majumder, Nathan Kallus, and Julian Mcauley.
\newblock Large language models as zero-shot conversational recommenders.
\newblock In \emph{Proceedings of the 32nd ACM International Conference on Information and Knowledge Management}, CIKM ’23. ACM, October 2023.
\newblock \doi{10.1145/3583780.3614949}.
\newblock URL \url{http://dx.doi.org/10.1145/3583780.3614949}.

\bibitem[Kojima et~al.(2022)Kojima, Gu, Reid, Matsuo, and Iwasawa]{NEURIPS20228bb0d291}
Takeshi Kojima, Shixiang~(Shane) Gu, Machel Reid, Yutaka Matsuo, and Yusuke Iwasawa.
\newblock Large language models are zero-shot reasoners.
\newblock In S.~Koyejo, S.~Mohamed, A.~Agarwal, D.~Belgrave, K.~Cho, and A.~Oh, editors, \emph{Advances in Neural Information Processing Systems}, volume~35, pages 22199--22213. Curran Associates, Inc., 2022.
\newblock URL \url{https://proceedings.neurips.cc/paper_files/paper/2022/file/8bb0d291acd4acf06ef112099c16f326-Paper-Conference.pdf}.

\bibitem[Zhang et~al.(2023)Zhang, Zhang, Li, and Smola]{zhang2023automatic}
Zhuosheng Zhang, Aston Zhang, Mu~Li, and Alex Smola.
\newblock Automatic chain of thought prompting in large language models.
\newblock In \emph{The Eleventh International Conference on Learning Representations}, 2023.
\newblock URL \url{https://openreview.net/forum?id=5NTt8GFjUHkr}.

\bibitem[Yao et~al.(2023)Yao, Yu, Zhao, Shafran, Griffiths, Cao, and Narasimhan]{yao2023tree}
Shunyu Yao, Dian Yu, Jeffrey Zhao, Izhak Shafran, Thomas~L. Griffiths, Yuan Cao, and Karthik Narasimhan.
\newblock Tree of thoughts: Deliberate problem solving with large language models, 2023.

\bibitem[Dhuliawala et~al.(2023)Dhuliawala, Komeili, Xu, Raileanu, Li, Celikyilmaz, and Weston]{dhuliawala2023chainofverification}
Shehzaad Dhuliawala, Mojtaba Komeili, Jing Xu, Roberta Raileanu, Xian Li, Asli Celikyilmaz, and Jason Weston.
\newblock Chain-of-verification reduces hallucination in large language models, 2023.

\bibitem[Wang et~al.(2023)Wang, Wei, Schuurmans, Le, Chi, Narang, Chowdhery, and Zhou]{wang2023selfconsistency}
Xuezhi Wang, Jason Wei, Dale Schuurmans, Quoc~V Le, Ed~H. Chi, Sharan Narang, Aakanksha Chowdhery, and Denny Zhou.
\newblock Self-consistency improves chain of thought reasoning in language models.
\newblock In \emph{The Eleventh International Conference on Learning Representations}, 2023.
\newblock URL \url{https://openreview.net/forum?id=1PL1NIMMrw}.

\bibitem[Schick et~al.(2023)Schick, Dwivedi-Yu, Dessì, Raileanu, Lomeli, Zettlemoyer, Cancedda, and Scialom]{schick2023toolformer}
Timo Schick, Jane Dwivedi-Yu, Roberto Dessì, Roberta Raileanu, Maria Lomeli, Luke Zettlemoyer, Nicola Cancedda, and Thomas Scialom.
\newblock Toolformer: Language models can teach themselves to use tools, 2023.

\bibitem[Chase(2022)]{ChaseLangChain2022}
Harrison Chase.
\newblock {LangChain}.
\newblock October 2022.
\newblock URL \url{https://github.com/hwchase17/langchain}.

\bibitem[Prasad et~al.(2023)Prasad, Koller, Hartmann, Clark, Sabharwal, Bansal, and Khot]{prasad2023adapt}
Archiki Prasad, Alexander Koller, Mareike Hartmann, Peter Clark, Ashish Sabharwal, Mohit Bansal, and Tushar Khot.
\newblock Adapt: As-needed decomposition and planning with language models, 2023.

\bibitem[Wu et~al.(2023{\natexlab{a}})Wu, Bansal, Zhang, Wu, Li, Zhu, Jiang, Zhang, Zhang, Liu, Awadallah, White, Burger, and Wang]{wu2023autogen}
Qingyun Wu, Gagan Bansal, Jieyu Zhang, Yiran Wu, Beibin Li, Erkang Zhu, Li~Jiang, Xiaoyun Zhang, Shaokun Zhang, Jiale Liu, Ahmed~Hassan Awadallah, Ryen~W White, Doug Burger, and Chi Wang.
\newblock Autogen: Enabling next-gen llm applications via multi-agent conversation, 2023{\natexlab{a}}.

\bibitem[Nori et~al.(2023)Nori, Lee, Zhang, Carignan, Edgar, Fusi, King, Larson, Li, Liu, Luo, McKinney, Ness, Poon, Qin, Usuyama, White, and Horvitz]{nori2023generalist}
Harsha Nori, Yin~Tat Lee, Sheng Zhang, Dean Carignan, Richard Edgar, Nicolo Fusi, Nicholas King, Jonathan Larson, Yuanzhi Li, Weishung Liu, Renqian Luo, Scott~Mayer McKinney, Robert~Osazuwa Ness, Hoifung Poon, Tao Qin, Naoto Usuyama, Chris White, and Eric Horvitz.
\newblock Can generalist foundation models outcompete special-purpose tuning? case study in medicine, 2023.

\bibitem[Nakano et~al.(2022)Nakano, Hilton, Balaji, Wu, Ouyang, Kim, Hesse, Jain, Kosaraju, Saunders, Jiang, Cobbe, Eloundou, Krueger, Button, Knight, Chess, and Schulman]{nakano2022webgpt}
Reiichiro Nakano, Jacob Hilton, Suchir Balaji, Jeff Wu, Long Ouyang, Christina Kim, Christopher Hesse, Shantanu Jain, Vineet Kosaraju, William Saunders, Xu~Jiang, Karl Cobbe, Tyna Eloundou, Gretchen Krueger, Kevin Button, Matthew Knight, Benjamin Chess, and John Schulman.
\newblock Webgpt: Browser-assisted question-answering with human feedback, 2022.

\bibitem[Liu et~al.(2023)Liu, Lai, Yu, Xu, Zeng, Du, Zhang, Dong, and Tang]{10.1145/3580305.3599931}
Xiao Liu, Hanyu Lai, Hao Yu, Yifan Xu, Aohan Zeng, Zhengxiao Du, Peng Zhang, Yuxiao Dong, and Jie Tang.
\newblock Webglm: Towards an efficient web-enhanced question answering system with human preferences.
\newblock In \emph{Proceedings of the 29th ACM SIGKDD Conference on Knowledge Discovery and Data Mining}, KDD '23, page 4549–4560, New York, NY, USA, 2023. Association for Computing Machinery.
\newblock ISBN 9798400701030.
\newblock \doi{10.1145/3580305.3599931}.
\newblock URL \url{https://doi.org/10.1145/3580305.3599931}.

\bibitem[Li et~al.(2023)Li, Zhu, Li, Yin, Sun, and Qiu]{li2023llatrieval}
Xiaonan Li, Changtai Zhu, Linyang Li, Zhangyue Yin, Tianxiang Sun, and Xipeng Qiu.
\newblock Llatrieval: Llm-verified retrieval for verifiable generation, 2023.

\bibitem[Wu et~al.(2023{\natexlab{b}})Wu, Zhang, and Huang]{wu-etal-2023-chain}
Dingjun Wu, Jing Zhang, and Xinmei Huang.
\newblock Chain of thought prompting elicits knowledge augmentation.
\newblock In Anna Rogers, Jordan Boyd-Graber, and Naoaki Okazaki, editors, \emph{Findings of the Association for Computational Linguistics: ACL 2023}, pages 6519--6534, Toronto, Canada, July 2023{\natexlab{b}}. Association for Computational Linguistics.
\newblock \doi{10.18653/v1/2023.findings-acl.408}.
\newblock URL \url{https://aclanthology.org/2023.findings-acl.408}.

\bibitem[Besta et~al.(2023)Besta, Blach, Kubicek, Gerstenberger, Gianinazzi, Gajda, Lehmann, Podstawski, Niewiadomski, Nyczyk, et~al.]{besta2023graph}
Maciej Besta, Nils Blach, Ales Kubicek, Robert Gerstenberger, Lukas Gianinazzi, Joanna Gajda, Tomasz Lehmann, Michal Podstawski, Hubert Niewiadomski, Piotr Nyczyk, et~al.
\newblock Graph of thoughts: Solving elaborate problems with large language models.
\newblock \emph{arXiv preprint arXiv:2308.09687}, 2023.

\bibitem[de~FSM~Russo and Camanho(2015)]{de2015criteria}
Rosaria de~FSM~Russo and Roberto Camanho.
\newblock Criteria in ahp: A systematic review of literature.
\newblock \emph{Procedia Computer Science}, 55:\penalty0 1123--1132, 2015.

\end{thebibliography}

\newpage
\appendix

\section{Prompt Design}
\label{sec:promt_design}

This section outlines the critical design considerations for selecting prompts essential for artifact generation. Our methodology involves precise fine-tuning of both the system and user prompts. The \textit{system prompt} is tailored to define the role of the Large Language Model (LLM), while the \textit{user prompt} focuses on task specification. 

\begin{figure}[h!]
\centering
\includegraphics[width=12.5cm]{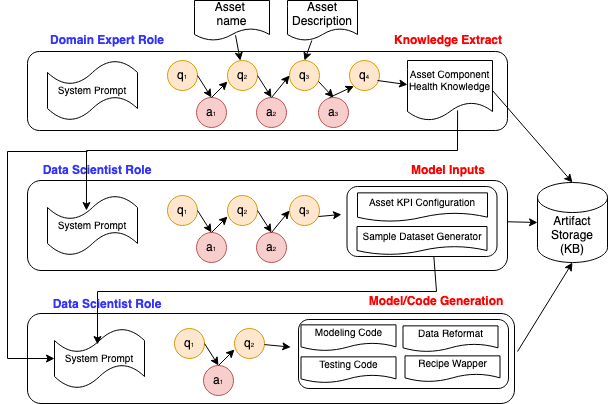}
\caption{Chained Pipelines for Decomposing Higher Level task into Smaller and Actionable Prompts for Artifact Generation}\label{fig:Pipelines for Artifact Generation}
\end{figure}

\subsection{System Prompt for Different Roles}
\vspace{10pt}

Our solution features a variety of roles, such as domain expert, data scientist or system integration architect, each carefully emulated by the LLM. In different situations, we formulate unique \textit{system prompt}s and \textit{user prompt}s to guide the LLM precisely. This specialized approach ensures the generation of diverse artifacts (shown in Figure~\ref{fig:Pipelines for Artifact Generation}), each tailored to meet the needs of the specific role and objectives of the session. For instance, consider the example of a \textit{system prompt} designed for extracting knowledge pertinent to estimating the health score from an asset's component health conditions. 

\begin{mybox}{System Prompt A. Domain Expert for Knowledge Extraction}
You are a helpful, respectful, and honest assistant, serving as a domain expert in industrial asset management. Your task is to provide specialized knowledge for the specific industrial asset. Ensure that your responses are safe, socially unbiased, and positive. Refrain from including any specific company details or contact information.

We seek detailed insights into the factors that affect industrial asset health, particularly to calculate the asset health score. Focus on assets such as turbines, electrical transformers, and compressors. The crucial aspects that influence asset health are generally grouped into three categories: a) the quality and health of asset components, b) the history of maintenance, failures, and repairs, and c) b) the profile of the asset, such as installation, asset location, and its age. In this session, our emphasis will be on understanding how the quality and health of asset components impact the overall asset score.
\end{mybox}

The following \textit{system prompt} generates the artifacts except the knowledge document. This prompt specifies the role of the LLM changed to a data scientist role. 

\begin{mybox}{Prompt B. Data Scientist}
You will play a role as the data scientist and software engineer. The user will describe the guidelines for assessing the component quality of an asset type. You need help to generate the necessary asset configurations, sample dataset, python code, and model recipes. The following are the general guidelines for asset quality. $asset\_class$ and $asset\_description$ as domain knowledge. 
\end{mybox}

\subsection{Prompts for Artifacts Generation}
\vspace{10pt}

We present examples of several \textit{user prompts}. The expected responses from LLM to these prompts constitute the artifacts that form the basis of our solution recipe.  The example used mainly comes from the asset health score scenario.  For the following prompts, the asset class and its description are replaced with a specific asset class at the time of execution. 

\subsubsection{Knowledge Document Generate}
\vspace{10pt}\label{subsec:knowledge_gen}

We have a series of Q\&As to help generate the final knowledge document. Using the health score estimation, we have for the last question from the user.

\begin{userpromptbox}{User Prompt: Knowledge Extraction}
Please help me export a markdown output as guidelines for analyzing the quality of asset component health or quality to overall asset health, the output has three parts:
    \begin{enumerate}
    \item It is the beginning part of the document, please briefly introduce the $asset\_class$, its usage in the
    various application and also include the introduction of its component,
    \item The factors that indicate the quality such as the deterioration of the $asset\_class$'s component quality;
    \item Highlights the sensors deployed or being able to use measure such components quality or its quality deterioration.  Please output as a list, each list contains a) the quality problem being monitored; b) the sensor used; and c) the reason of using such sensor.
    \end{enumerate}

\end{userpromptbox}

\subsubsection{Asset Configuration Generate}
\vspace{10pt}

For health score estimates from the health component, we utilize two distinct approaches: the analytic hierarchical process (AHP) and a weighted approach. These methodologies will be integrated into the system configuration, which specific customers can tailor during deployment.

The AHP method~\cite{de2015criteria} is particularly advantageous due to its user-friendly configuration. It is designed to be more easily adjustable by customers or end-users, making it a preferable choice for its accessibility and adaptability in diverse application settings. 

There are two configuration files generated.  The first one defines the health indicator for each sensor.  The second one defines the parameters required for aggregating the sensor or measurement values for the final health score. 

\begin{userpromptbox}{User Prompt: Asset Health Indicator Configuration Generation}
Let us assume that we have a measurement of the results of sensor output and assign a value as [poor, medium, good, excellent].  Can you give me a configuration with the thresholds (please specify the physical unit of defining poor, medium, good, and excellent) as health indicators for the given sensor in YAML format?   We should have results outputted as the sequence of sensors.  For each sensor, provide the ranges for all [poor, medium, good, excellent]
\end{userpromptbox}

\begin{userpromptbox}{User Prompt: Asset Health Weighted and AHP Approach Parameter Generation}
Great. Now, I need a configuration file that accomplishes two main objectives. First, it should define the relative scores for sensor categories [poor, medium, good, excellent], ensuring each category is assigned a value within the range of [0, 1]. Second, the file should detail two approaches for aggregating the health score into a scale of [0, 100]. The first approach will utilize a weighted system; please provide the relative weights for each factor as defined by the sensor, with the total sum of weights equaling 1.0. The second approach should be based on the Analytical Hierarchical Process (AHP), requiring the inclusion of pairwise comparisons to determine the relative importance of each factor. Finally, remember to exclude the asset type name from the YAML file.
\end{userpromptbox}

\begin{userpromptbox}{User Prompt: Aggregation Configuration Generation}
I want to have a configuration file defined:
\begin{enumerate}
    \item The relative score for the scale of sensors [poor, medium, good, excellent]
    \item Two approaches to aggregate the health score into [0, 100].  The first approach uses the weights approach; please give me the relative weight of the sensor-defined factor.   The sum of the weight should be 1.0.  The second is the analytical hierarchical process (APH) process.  I need to have the pairwise relative importance.
\end{enumerate}
\end{userpromptbox}

\subsubsection{Sample Dataset Generate}
\vspace{10pt}

The generation of the sample dataset is a two-step process. Initially, a concise dataset is created based on the information within the knowledge document. Following this, the initial dataset serves as a template to produce simulated code, which can then be used to generate synthetic data of any desired size.

This is an example of the sample dataset generation for health score estimated using component quality. 

\begin{userpromptbox}{User prompt (Data Scientist): Small Set of Sample Data Generate}
Great. Let us generate a sample asset for component quality to estimate the health score with $num\_assets$ assets.  The row of the file (say, using the CSV form) contains the asset ID,  sensor name, value, the time of the measurement, and physical unit if relevant.  The CSV file should align with the above configuration file for asset health.  Ensure we complete the $num_assets$ assets generated without missing any sensor values.  Do not contain the code to transfer the dataset to have the asset score. 
\end{userpromptbox}

\begin{userpromptbox}{User Prompt (Data Scientist): Synthetic Code Generation}
Please utilize the sample data generated above and help develop a Python code to create synthetic data for any number of assets for health score estimation.  Ensure the output from the synthetic data simulator aligned with the sample data structure.  
\end{userpromptbox}

\subsubsection{Model Generate for KPI of a Given asset class}
\vspace{10pt}

The model has to be aligned to the scikit learn.  We have both the engineering and ML models.  The guideline for generation model with user prompt. 

\begin{userpromptbox}{User Prompt: Model Generation}
As a data scientist, your task is to develop a unified Python class that standardizes the interface for various models, treating them as functions of the form $f\left(x\right) = y$. This class should extend Scikit-Learn's BaseEstimator and TransformerMixin, handling both a configuration file and a dataset dataframe. It must include essential methods like fit() and predict(), ensuring compatibility with Scikit-Learn's ecosystem. Your implementation should be flexible to support different model types and include clear documentation for ease of use.

We use the health score range confirmation to conversion factor, the AHP configuration for the overall score calculation.  Then, we need to use the sample dataset to have the conversion.  Please generate the Python code for us. 
\end{userpromptbox}

\begin{userpromptbox}{User Prompt (Data Scientist): Model Factory Wrapper}
We need to generate the Python class that takes the configuration file and the dataset, passes them into the model class generated above, and tracks the process in an MLFlow server. 
\end{userpromptbox}

\subsection{Improvement of Knowledge Extraction with Iterative Prompting}\label{sec:appendix_enhancement}
\vspace{10pt}

The following system and user prompt technique applied to help to improve the quality of knowledge extraction. We simple iterative ask the LLM to repeatedly generate the same answer with the requirements of improvement itself quality.  In following example, we only display one example (one-shot) due to the limitation of the paper, however, in the experiment, we choose two-shot approach in the system prompt. We also ask the output coming out the confidence level as part of outputs. However, we felt that this confidence level might not be truly reflects the confidence of the context generated rather than a hallucination.

\begin{mybox}{System Prompt: Domain Expert in for Knowledge Extraction}
You are a helpful AI assistant. Given an industrial asset, generate its description and its components. 

Also generate confidence score of answer. The confidence score should be based on the level of detail and accuracy of the generated description, and is subjective to the knowledge and understanding of the system generating the response. Do not provide note and explanation to the confidence score. Do not generate any information other than the description and confidence score. You can take a look at the following examples. Output should contain only generate Answer and Confidence score. 

Question: Describe Wind Turbine Asset.

Answer: A wind turbine is a tall tower with long, spinning blades that capture the wind's energy.  The wind pushes the blades around, which turns a shaft that connects to a generator. The generator converts the mechanical energy of the spinning shaft into electricity that we can use in our homes and businesses. 
The main components of a wind turbine include the rotor blades, the hub, the main shaft, the gearbox, the generator, the yaw system, and the control system.
Confidence: \textbf{90\% (TOKENSTOP)}
\end{mybox}

To enhance the output document to ensure accuracy, consistency, and completeness.  We experiment with the reinforcement approach for the user prompts by initialing the prompt as follows. Then, repeat the second user prompt for $N$ times (say $N=10$). 

\begin{userpromptbox}{User Prompt: Initial Prompt}
Describe $asset\_class$, and its components.
\end{userpromptbox}

\begin{userpromptbox}{User Prompt: Repeatable Prompt - $N$ Times}
Describe $asset\_class$, and its components. Generate better solutions in terms of readability, accuracy, and completeness.
\end{userpromptbox}

We have a comprehensive experiment using this technique to improve knowledge extraction.  The detailed results can be found in Section~\ref{subsec:knowledge_extraction_experiment}.
\section{KPI-centric Taxonomy to Prompt Auto-Generation}
\label{kctax}
In this section, we are showing examples for how KPITaxo2Prompt auto-generates prompt/step sequences with various pre-defined KPI taxonomy using LLM. The prompt consists of \textbf{System Prompt} and \textbf{User Prompt}. In System Prompt shown in Table \ref{tab:kpitaxo2prompt}, we are providing context, the taxonomy descriptions, and one goal along with its Think-Step-ExecOrder answer. The goal is considered as one-shot example. 

Next, in User Prompt 1, we demonstrate how LLM responds to a new goal where its KPI, asset health, is identical, but a different method is requested for analysis. In User Prompt 2, we demonstrate how LLM responds to a new goal where its KPI, asset sustainability in this case, differs from the one-shot example. Note that the user prompt is independent of asset types. We replace asset class and description with specific strings in the downstream tasks.

\begin{figure}[h]
\centering
\includegraphics[width=12cm]{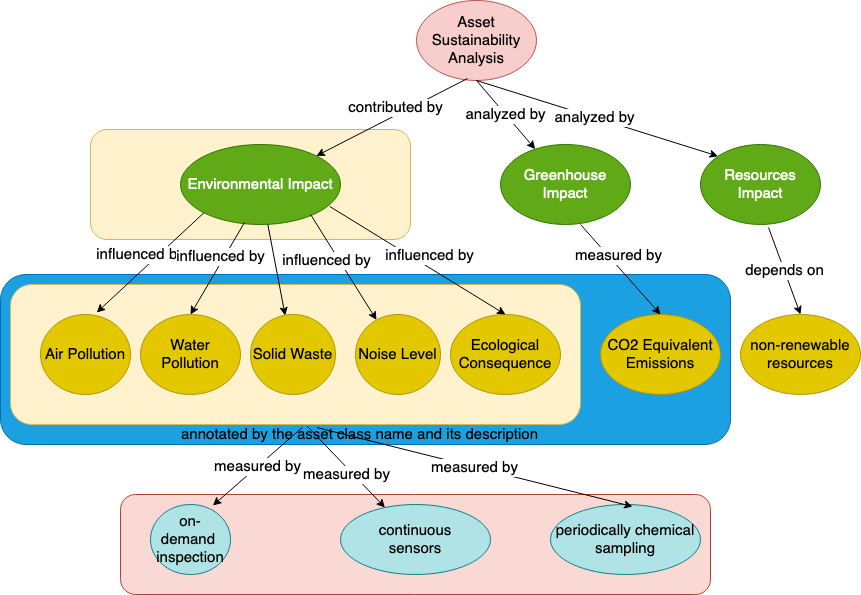}
\caption{KPI-centric Taxonomy for Sustainability (Knowledge Document Generation)}\label{fig:kpitree_sus}
\end{figure}

\begin{mybox}{Materialized Taxonomy Description - Asset Health}
\label{prompt:kpitaxo}
asset health taxonomy:\\
asset health is root node\\
asset health is analyzed by component quality\\
asset health is analyzed by historical record\\
asset health is analyzed by asset profile\\
component quality is impacted by mechanical issue\\
component quality is impacted by electrical issue\\
component quality is impacted by thermal health issue\\
component quality is impacted by chemical health issue\\
mechanical issue is measured by on-demand inspection\\
mechanical issue is measured by continuous sensors\\
mechanical issue is measured by periodic chemical sampling\\
electrical issue is measured by insulation\\
historical record is source by asset maintenance\\
historical record is source by failure\\
historical record is source by repair history\\
asset profile is recorded by age\\
asset profile is recorded by operating hours\\
asset profile is recorded by idle hours
\end{mybox}

\begin{mybox}{Materialized Taxonomy Description - Asset Sustainability}
\label{prompt:kpitaxo_sus}
asset sustainability taxonomy:\\
asset sustainability is root node\\
asset sustainability is analyzed by greenhouse impact\\
asset sustainability is analyzed by natural resources impact\\
asset sustainability is contributed by environmental impact\\
environmental impact is influenced by air pollution\\
environmental impact is influenced by water pollution\\
environmental impact is influenced by solid waste\\
environmental impact is influenced by noise level\\
environmental impact is influenced by ecological consequence\\
greenhouse impact is measured by co2-equivalent emissions\\
resources impact depends on non-renewable resources\\
\end{mybox}

\begin{smallerfont}

\onecolumn
\begin{table}[htbp]
    \centering
    \caption{System Prompt: Prompt Generation with LLM and Taxonomy}
    \label{tab:kpitaxo2prompt}
    \begin{tabular}{@{}p{16cm}@{}}
        \toprule
        \textbf{KPITaxo2Prompt: System Prompt} \\
        \midrule
Your task is to develop a short plan to help me accomplish my goal using the provided taxonomy in a couple of steps. You can take the help of the taxonomy below to create a new plan.
Taxonomy can be understood as follows: [parent] is [relation] by [child], where parent and child are nodes in taxonomy, and relation is the connection between parent and child nodes. Taxonomy is non-cyclic, it means same node cannot be connected to itself. You are given an agent
that recursively can produce a plan by traversing the given taxonomy.
Here is the asset health taxonomy. \\
<insert asset health taxonomy description here. Read Materialized Taxonomy Description for reference.> \\
Here is the asset sustainability taxonomy: \\
<insert asset sustainability taxonomy description here. Read Materialized Taxonomy Description for reference.> \\
\\
\textbf{Goal}: calculate asset health using the component health condition. \\
\# Think: My target is component. I will use asset health taxonomy to devise a plan for component condition-based health score estimation.
From the asset health taxonomy, four options start with the node component. Mechanical issues, electrical issues, thermal health issues, and chemical health issues impact the components. 
Subsequently, the mechanical issue is measured by on-demand inspection, continuous sensors, periodic chemical sampling, and so on. I should prepare the plan using the given taxonomy. I will traverse the taxonomy in a top-down fashion starting from the component node. 
   
\textbf{Step 1}: Let us focus on the component-based asset health. We are interested in the factors coming from the 1. Mechanical, 2. Electrical, 3. Thermal, and 4. Chemical quality issues for the asset health. Can you give me detailed   guidelines for identifying factors impacting overall asset health?

\# Think: Now, I will traverse taxonomy for each child node connected to component node and so on. From the asset health taxonomy, component is impacted by mechanical issue and thus we will visit mechanical issue node to identify factors associated with mechanical issue.
I know that I cannot use a partial recipe.

\textbf{Step 2}: Let us focus on mechanical issue of component. We are  interested in the factors coming from the 1. on-demand inspection, 2. continuous sensors,  and 3. periodic chemical sampling mechanic for the asset health. Can you give me detailed guidelines for identifying factors impacting overall asset health?

\# Think: Now that I have visited mechanical issue node, I will focus on other child node connected to component node. From the asset health taxonomy, component is also impacted by electrical issue, thermal health issue and chemical health issue. I know that I cannot use a partial recipe. So my goal is not satisfied, I need to visit more nodes by repeating Step 2 three more times.
 
\textbf{Step 3}: Let us focus on electrical issue of component. We are interested in the factors coming from the 1. insulin for the asset health. Can you give me detailed guidelines for identifying factors impacting overall asset health?
    
\textbf{Step 4}: Let us focus on thermal health issue of component. We are interested in the factors coming from the thermal health issue of component for the asset health. Can you give me detailed guidelines for identifying factors impacting overall asset health?

\textbf{Step 5}: Let us focus on chemical health issue of component. We are interested in the factors coming from the chemical health issue of component for the asset health. Can you give me detailed guidelines for identifying factors impacting overall asset health?
        
\# Think: We have visited all four nodes: mechanical issue, electrical issue, thermal health issue and chemical health issue. 
Now that I have traversed all the child and sub-child node of component node. I should repeat Step 2 for any other node that is not visited so far. I know that I cannot use a partial recipe.

\textbf{Step 6}: Collect all the discovered factors so far.  

\# Think: We now shift our focus to discovered factors, asset class, and asset description. 

\textbf{Step 7}: Let us assume that I am only interested in evaluating the asset health based on the currently available data and discovered {factors}. I need your help to identify the factors that indicate the deterioration of the ${asset\_class}$'s component quality,  and those factors should be able to be monitored and quantified in future data.

\textbf{Step 8}: Great, this answer is an excellent general guideline. Let's focus on a  specific asset type: ${asset\_class}$ to analyze the component health on the deterioration of the ${asset\_class}$'s health score.  Here is a detailed explanation: ${asset\_description}$. Do not include any specific company contact information.

\textbf{Step 9}: We need to be very specific for asset type ${asset_class}$. Here is a detailed explanation: ${asset\_description}$. Let us focus on the factor, Key Performance Indicators (KPIs) and reports. Please list all of them with the factors, KPIs and report, explanation, and impact on asset health.

\textbf{Step 10}: Generate a Python code that takes the component as an input node from taxonomy and returns "yes it is there" if any child node of the input node has a word sensor into their name otherwise, it returns skip. Code will traverse the taxonomy in a top-down fashion starting from the input node. Please handle KeyError, especially before the dictionary key. Use the dictionary to store the parent and child relationship of taxonomy.  
A leaf node should also be present in the dictionary with an empty child. Only generate Python code in PEP standard, 
Do not generate an explanation. Generate a single Python function as an entry point call me(node='component'). If you have any explanations, please put them into Python code.

\textbf{Step 11}: Please help me export a markdown output as guidelines for analyzing the quality of asset component health or quality to overall asset health, the output has three parts where part 1 and part 2 are compulsory and part 3 is needed if it is relevant. \\
Part 1:  It is the beginning part of the document, please briefly introduce the ${asset\_class}$, its usage in the various application and also include the introduction of its component \\
Part 2: the factors/KPIs that indicate or impact the component-based asset health using the categories of mechanical issue, electrical issue, thermal health issue and chemical health issue. \\
Part 3 (optional): highlights the sensors deployed or being able to use measure such components quality or its quality deterioration.  Please output as a list, each list contains a) the quality problem being monitored; b) the sensor used, and c) the reason of using such sensor. 
\# Think: To succeed, I need to perform all these steps, one after the other. So, I need to use the "AND" operator. \\
Execution Order: (Step 1 AND Step 2 AND Step 3 AND Step 4 AND Step 5 AND Step 6 AND Step 7 AND Step 8 AND Step 9 AND Step 10 AND Step 11) Goal completed!

Here is a different goal with different taxonomy relations. Your task is to develop a short plan to help me accomplish my goal in a couple of steps using the provided taxonomy. You can take the help of the asset health taxonomy above to create a new plan. Keep in mind that:
(1) It is okay to update the asset health taxonomy to the original.
(2) Be very careful with the adding of duplicate node in asset health taxonomy. (3) You cannot use a partial asset health taxonomy. (4) You cannot repeat the same step in the plan. (5) Do not traverse newly created nodes in taxonomy.\\ 
\bottomrule
\end{tabular}
\end{table}
\end{smallerfont}

\begin{mybox}{KPITaxo2Prompt: User Prompt 1}
Goal: Analyze the asset health using asset profile information
\end{mybox}
\begin{table}[htbp]
    \centering
    \begin{tabular}{@{}p{16cm}@{}}
        \toprule
        \textbf{KPITaxo2Prompt LLM Response 1: Generated Steps for the Asset Health Analysis using Asset Profile} \\
        \midrule
\textbf{Step 1}: Let us focus on asset profile based asset health. We are
    interested in the factors coming from the 1. age, 2. operating hours, and 3. idle hours for the asset health. Can you give me detailed
    guidelines for identifying factors impacting overall asset health?\\
\textbf{Step 2}: Let us focus on age of asset profile. We are
    interested in the factors coming from the 1. year for the asset health. Can you give me detailed
    guidelines for identifying factors impacting overall asset health?\\
\textbf{Step 3}: Let us focus on operating hours of asset profile. We are
    interested in the factors coming from the 1. hour for the asset health. Can you give me detailed
    guidelines for identifying factors impacting overall asset health?\\
\textbf{Step 4}: Let us focus on idle hours of asset profile. We are
    interested in the factors coming from the 1. hour for the asset health. Can you give me detailed
    guidelines for identifying factors impacting overall asset health?\\
\textbf{Step 5}: Collect all the discovered factors so far.\\
\textbf{Step 6}: Let us assume that I am only interested in evaluating the asset health
    based on the currently available data and discovered factors. I need your help to identify
    the factors that indicate the deterioration of the ${asset\_class}$'s asset profile quality,
    and those factors should be able to be monitored and quantified in the future
    data.\\
\textbf{Step 7}: Great, this answer is an excellent general guideline. Let's focus on a
    specific asset type: ${asset\_class}$ to analyze the asset profile health on the
    deterioration of the ${asset\_class}$'s health score.  
    Do not include any specific company contact information.\\
\textbf{Step 8}: We need to be very specific for asset type ${asset\_class}$. Here is a detailed
    explanation: ${asset\_description}$. Let us focus on the factor, Key Performance Indicators (KPIs)
    and reports. Please list all of them with the factors, KPIs and report,
    explanation, and impact on asset health.\\
\textbf{Step 9}: Please help me export a markdown output as guidelines for analyzing the quality of
    asset profile health or quality to overall asset health, the output has three parts where part 1 and part 2 are compulsory and part 3 is needed if it is relevant.\\
    Part 1:  It is the beginning part of the document, please briefly introduce the ${asset\_class}$, its usage in the
    various application and also include the introduction of its component\\
    Part 2: the factors/KPIs that indicate or impact the asset profile-based asset health using the categories of age, operating hours and idle hours\\
    Part 3 (optional): highlights the sensors deployed or being able to use measure such asset profile quality
    or its quality deterioration.  Please output as a list, each list contains a) the quality problem being
    monitored; b) the sensor used, and c) the reason of using such sensor.\\
Execution Order: (Step 1 AND Step 2 AND Step 3 AND Step 4 AND Step 5 AND Step 6 AND Step 7 AND Step 8 AND Step 9) Goal completed!\\
\bottomrule
\end{tabular}
\end{table}
\newpage
\begin{mybox}{KPITaxo2Prompt: User Prompt 2 }
Goal: Analyze the Asset Sustainability from it Environmental Impact
\end{mybox}
    
\begin{table}[htbp]
    \centering
    \begin{tabular}{@{}p{16cm}@{}}
        \toprule
        \textbf{KPITaxo2Prompt LLM Response 2: Generated Steps for Sustainability Analysis of Asset based on its Environmental Impact} \\
        \midrule
\textbf{Step 1}: Let us focus on environmental impact based asset sustainability. We are
    interested in the factors coming from the 1. air pollution, 2. water pollution, 3. solid waste,
    4. noise level, and 5. ecological consequence for the asset sustainability. Can you give me detailed
    guidelines for identifying factors impacting overall asset sustainability?\\
    \textbf{Step 2}: Let us focus on air pollution of environmental impact. We are
    interested in the factors coming from the 1. co2-equivalent emissions for the asset sustainability. Can you give me detailed
    guidelines for identifying factors impacting overall asset sustainability?\\
    \textbf{Step 3}: 
    Let us focus on water pollution of environmental impact. We are
    interested in the factors coming from the 1. biochemical oxygen demand for the asset sustainability. Can you give me detailed
    guidelines for identifying factors impacting overall asset sustainability?\\
\textbf{Step 4}: 
    Let us focus on solid waste of environmental impact. We are
    interested in the factors coming from the 1. volume and 2. weight for the asset sustainability. Can you give me detailed
    guidelines for identifying factors impacting overall asset sustainability?\\
\textbf{Step 5}: 
    Let us focus on environmental impact based on noise level. We are
    interested in the factors coming from the 1. decibel for the asset sustainability. Can you give me detailed
    guidelines for identifying factors impacting overall asset sustainability?\\
\textbf{Step 6}: Let us focus on environmental impact based on ecological consequence. We are
    interested in the factors coming from the 1. species diversity for the asset sustainability. Can you give me detailed
    guidelines for identifying factors impacting overall asset sustainability?\\
\textbf{Step 7}: 
    Collect all the discovered factors so far.\\
\textbf{Step 8}: 
    Great, this answer is an excellent general guideline. Let's focus on a
    specific asset type: ${asset\_class}$ to analyze the environmental impact on the sustainability of the ${asset\_class}$. Do not include any specific company contact information.\\
\textbf{Step 9}: 
    We need to be very specific for asset type ${asset\_class}$. Here is a detailed
    explanation: ${asset\_description}$. Let us focus on the factor, Key Performance Indicators (KPIs)
    and reports. Please list all of them with the factors, KPIs and report,
    explanation, and impact on asset sustainability.\\
\textbf{Step 10}: 
    Please help me export a markdown output as guidelines for analyzing the quality of
    asset sustainability using environmental impact, the output has three parts where part 1 and part 2 are compulsory and part 3 is needed if it is relevant.\\
    Part 1:  It is the beginning part of the document, please briefly introduce the ${asset\_class}$, its usage in the
    various application and also include the introduction of its environmental impact\\
    Part 2: the factors/KPIs that indicate or impact the environmental impact using the categories of air pollution, water pollution, solid waste, noise level and ecological consequence\\
    Part 3 (optional): highlights the sensors deployed or being able to use measure such environmental impact
    or its quality deterioration.  Please output as a list, each list contains a) the environmental impact problem being
    monitored; b) the sensor used, and c) the reason of using such sensor.
    \\
    Execution Order: (Step 1 AND Step 2 AND Step 3 AND Step 4 AND Step 5 AND Step 6 AND Step 7 AND Step 8 AND Step 9 AND Step 10) Goal completed!\\

\bottomrule
\end{tabular}
\end{table}
\newpage

\section{PromptSequence Execution Engine Generation}
\label{pmq}
\begin{figure}[h]
\centering
\caption{Generated Knowledge Document for Sustainability Analysis (Example: Industrial Furnace)}
\vspace{10pt}
\frame{\includegraphics[width=15cm]{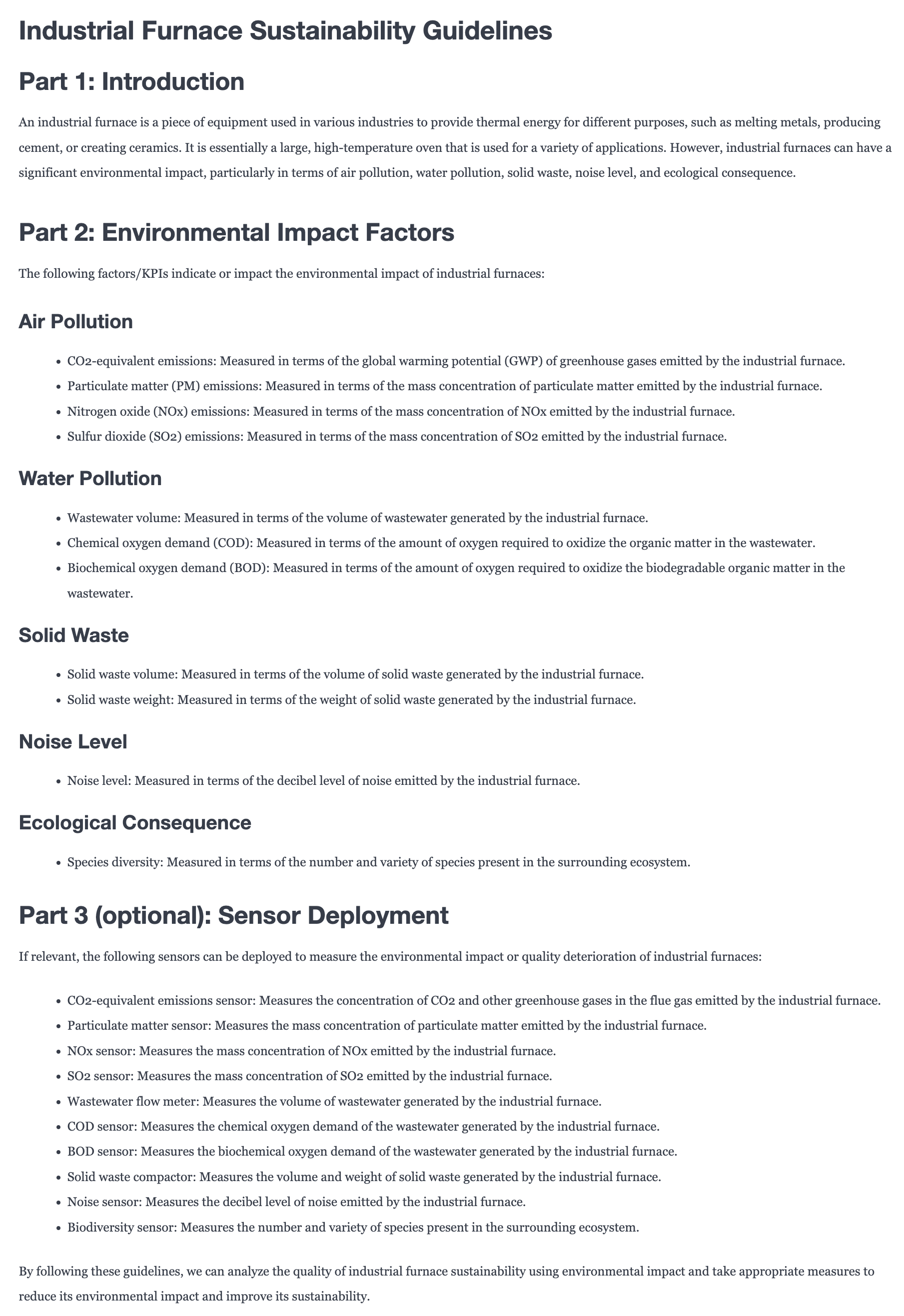}}

\label{fig:knowledgedoc}
\end{figure}

In this section, we showcase an example of knowledge document generated by PromptSequence Execution Engine. We fed the 10 steps from LLM Response 2 to the Engine, then replace the ${asset\_class}$ with \textit{Industrial Furnace} and ${asset\_description}$ with a concrete description of \textit{Industrial Furnace}. We choose All Questions with ReAct ($All_{QREACT}$) as the approach for constructing the following user prompt. Finally, the prompt generated the knowledge document shown in Figure \ref{fig:knowledgedoc}. And Figure \ref{fig:ref1} and \ref{fig:ref2} are the result document after reference generation.

\begin{mybox}{User Prompt for PromptSequence Execution Engine}
You don't have any tools to use. Please ignore this information: {tool\_names}, {tools}, {agent\_scratchpad}.

Now let's focus on asset class ${asset\_class}$. The asset description is ${asset\_description}$. Now we have prepared the following questions for the asset ${asset\_class}$.\\
<Insert the first 9 steps from User Prompt 2>\\
You must give answers to these questions one by one and you must think step by step using the following format:\\

Question: the input question you must answer\\
Thought: you should think carefully \\
Answer: the answer to the question. Keep it precise. Limit the answer length to 200 words.\\
... (this Question/Thought/Answer can repeat N times)\\

Thought: Now I know how to answer the final request: <insert the last step from User Prompt 2>\\
Final Answer: the final answer to the request. Please write a nice article summarizing the answers you have in the thought process.\\

Begin!

\end{mybox}

\begin{figure}[h]
\centering
\caption{Knowledge Document for Sustainability Analysis with Reference (Example: Industrial Furnace)}
\vspace{10pt}
\frame{\includegraphics[width=15cm]{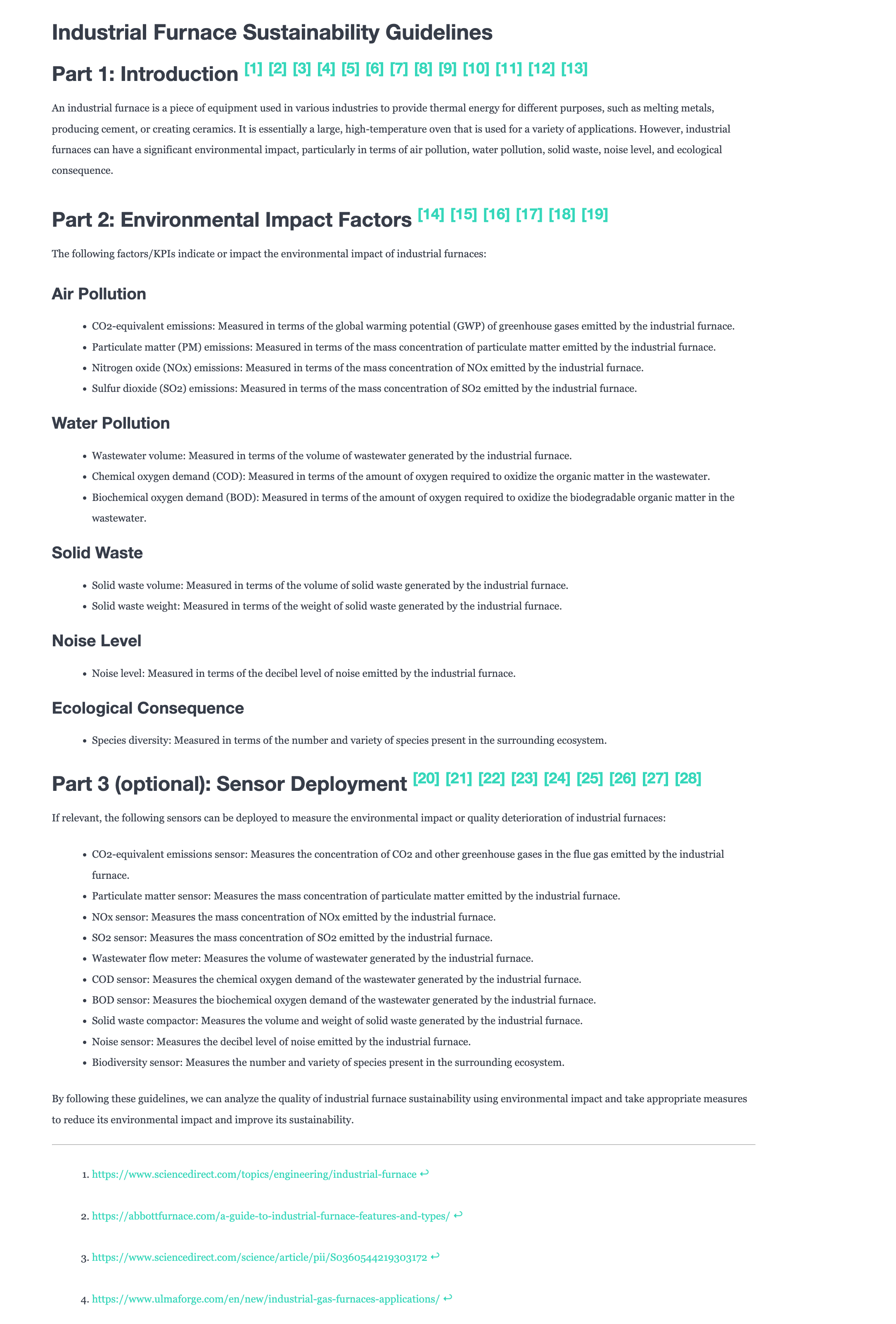}}

\label{fig:ref1}
\end{figure}
\begin{figure}[h]
\centering
\caption{Continued: Knowledge Document for Sustainability Analysis with Reference (Example: Industrial Furnace)}
\vspace{10pt}
\frame{\includegraphics[width=15cm]{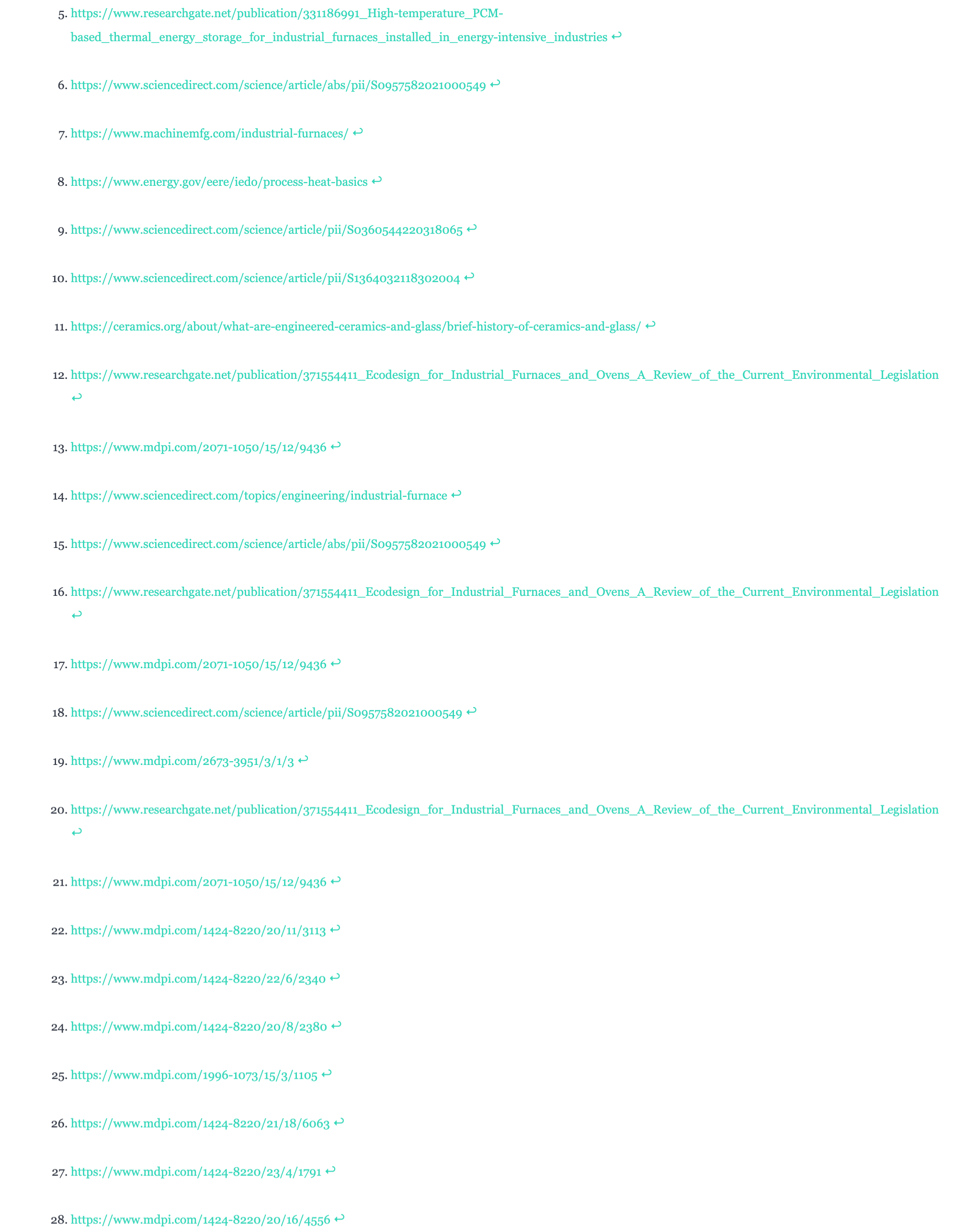}}

\label{fig:ref2}
\end{figure}

\end{document}